\documentclass[compsoc,conference,a4paper,10pt,times]{IEEEtran}
\IEEEoverridecommandlockouts
\usepackage{cite}
\usepackage[normalem]{ulem}
\useunder{\uline}{\ul}{}
\usepackage{amsmath,amssymb,amsfonts}
\usepackage{algorithmic}
\usepackage{graphicx}
\usepackage{textcomp}
\usepackage{bmpsize}
\usepackage{xcolor}
\usepackage{lipsum}
\usepackage{amsthm}
\usepackage{caption}
\usepackage{bbm}
\usepackage{wasysym}
\usepackage{booktabs}
\usepackage{multirow}
\usepackage{enumitem}
\usepackage{subcaption}
\usepackage{threeparttable} 
\usepackage[colorlinks=true,urlcolor=black]{hyperref}
\def\BibTeX{{\rm B\kern-.05em{\sc i\kern-.025em b}\kern-.08em
    T\kern-.1667em\lower.7ex\hbox{E}\kern-.125emX}}

\newtheorem{definition}{Definition}

\begin{document}

\title{SoK: Challenges in Tabular Membership Inference Attacks}


\author{
\IEEEauthorblockN{
{Cristina Pêra}\IEEEauthorrefmark{1}\IEEEauthorrefmark{4},
Tânia Carvalho\IEEEauthorrefmark{2},
Maxime Cordy\IEEEauthorrefmark{2},
Luís Antunes{\IEEEauthorrefmark{1}\IEEEauthorrefmark{3}}
}\\
\IEEEauthorblockA{
\IEEEauthorrefmark{1}\textit{Department of Computer Science, Faculty of Sciences of the University of Porto, Portugal}\\[0.3ex]
\IEEEauthorrefmark{2}\textit{SnT, University of Luxembourg, Luxembourg}\\[0.3ex]
\IEEEauthorrefmark{3}\textit{TekPrivacy, Portugal}\\[0.3ex]
\IEEEauthorrefmark{4}corresponding author: up201907321@up.pt
}
}

\maketitle

\begin{abstract}
Membership Inference Attacks (MIAs) are currently a dominant approach for evaluating privacy in machine learning applications. Despite their significance in identifying records belonging to the training dataset, several concerns remain unexplored, particularly with regard to tabular data. In this paper, first, we provide an extensive review and analysis of MIAs considering two main learning paradigms: centralized and federated learning. We extend and refine the taxonomy for both. Second, we demonstrate the efficacy of MIAs in tabular data using several attack strategies, also including defenses. Furthermore, in a federated learning scenario, we consider the threat posed by an outsider adversary, which is often neglected. Third, we demonstrate the high vulnerability of single-outs (records with a unique signature) to MIAs. Lastly, we explore how MIAs transfer across model architectures. Our results point towards a general poor performance of these attacks in tabular data which contrasts with previous state-of-the-art. 
Notably, even attacks with limited attack performance can still successfully expose a large portion of single-outs. Moreover, our findings suggest that using different surrogate models makes MIAs more effective.  

\end{abstract}

\section{Introduction}

In the past decade, Artificial Intelligence (AI) has reached astonishing progress. Currently, nearly every aspect of daily life involves at least one AI-driven system, for instance, personalized recommendations, virtual assistants, and autonomous vehicles. Although such technologies automate processes and greatly enhance convenience and efficiency, their rapid and often uncontrolled development has also raised societal concerns, in particular regarding privacy, as personal information is being stored and reused. Therefore, the European Parliament and Council created the AI Act, an European Union AI regulation~\cite{aiact}. The AI Act builds on and must operate in harmony with the GDPR~\cite{gdpr}. While GDPR governs how data are collected, stored, and processed, AI Act governs how AI systems are designed, developed, and deployed.


The broad spectrum of domains in which AI can be applied underscores the importance of developing tools and methodologies that can effectively support both developers and existing systems in evaluating their compliance.

A quantitative privacy measure that is currently on the spotlight is Membership Inference Attack (MIA)~\cite{Shokri-MIA-Against-ML,pyrgelis2017knock,yan2022membership}. This is, so far, the most widely used and easiest to interpret approach. Put simply, MIA determines whether a specific record (i.e. an individual) is part of the training dataset. For example, if an adversary determines that a medical record was used to train a cancer prediction model, they may infer that the individual has cancer. Therefore, MIAs can help prevent privacy breaches by revealing details about the training data of machine learning (ML) models. MIAs can also serve as an effective auditing tool for data privacy and compliance. For instance, under the legal requirement of the \textit{right to be forgotten} (Art. 17, GDPR), MIAs can verify if the model does not include the specific data point following a deletion request.

This type of attack has been explored on a wide variety of data types and learning tasks. Initially, MIAs were studied in the context of classification problems~\cite{li2021membership,Shokri-MIA-Against-ML,Hierarchical-tree-based,MIA-first-prin,Salem-ML-Leaks,rmia,Enhanced-MIA-Game}. Since then, research has been extended to other domains such as regression~\cite{gupta2021membership}, recommendation systems~\cite{zhang2021membership,chi2024shadow,he2025membership}, graph learning~\cite{wu2021adapting}, natural language processing~\cite{shejwalkar2021membershipnlp}, and even wireless signal analysis~\cite{shi2022membership} and location data~\cite{pyrgelis2017knock}. Recently, attention has shifted towards MIAs in Large Language Models (LLMs)~\cite{meeus2025sok,wu2025membership,he2025towards,wen2024membership}. 

All these examples are conducted within a centralized paradigm, where data are stored and processed on a single machine or server, giving the model full access to the complete dataset during training. Another important guideline outlined in data privacy regulations concerns the sharing of data between organizations. To address the challenges associated with secure sharing data, Federated Learning (FL) has emerged as a promising solution~\cite{mcmahan2017communication,zhang2021survey, wen2023survey}.
FL is a distributed ML paradigm that enables multiple participants to collaboratively train a ML model (global model) without directly sharing their private data. This distributed paradigm has been relevant in domains such as healthcare~\cite{nguyen2022federated, pfitzner2021federated} and finance~\cite{liu2023efficient,chatterjee2023federated}.
Despite reducing the threats associated with data sharing, FL is still vulnerable to MIAs~\cite{central-fl-comp,wang2023gbmia,he2024enhance,zhao2021user}. 

Although multiple data domains have been explored for MIAs, tabular data remains overlooked. Tabular data are the predominant data format in real-world applications. Sectors such as healthcare, finance, demographics and government statistics depend heavily on structured data. 
Nonetheless, the nature of attributes and the presence of sensitive information in tabular data make privacy analysis essential. Yet, most existing MIA evaluations neglect this setting, in particular simple tabular datasets.

In this paper, we systematically analyze tabular MIAs in two different learning paradigms. First, we explore such attacks in centralized learning. Then, we delve into distributed learning, i.e., FL, to understand to what extend FL can be a defense against MIAs. 
We focus our analysis on MIAs in supervised learning, in particular classification tasks, giving the broad applicability in real scenarios.

We identify four main challenges that remain unaddressed. First, models trained on simple tabular data are rarely evaluated, making MIAs less reliable in such data domain. Second, FL which aims to safeguard the confidentiality of the training data, is commonly attacked from a participants' side, which requires additional background knowledge. 
Third, single outs, referring to isolated individuals present in microdata (individual-level data), have received no attention in the MIA literature, despite their clear implications for privacy breaches. Lastly, fewer studies have explored model transferability under limited knowledge, where the intruder uses mismatched architectures, a gap that is even more pronounced for tabular MIAs. 
\textbf{Overall, we make the following contributions.}

\begin{enumerate}[label=\roman*)]
    \item We thoroughly investigate existing research on both centralized and FL concerning MIAs and respective defenses for each learning paradigm. 
    Several researchers have surveyed MIAs. However, the majority focuses on centralized learning~\cite{niu2024survey,hu2023defenses,niu2025comparing,meeus2025sok} and,
    although some prior works summarize MIAs including FL~\cite{bai2024membership,rao2024privacy}, they lack a comprehensive systematization of the literature. For instance, MIAs in centralized learning have not been extensively analyzed~\cite{bai2024membership}, and given the exploration of general privacy threats, MIAs receive little attention~\cite{rao2024privacy}.
    We fill this gap by concentrating our efforts on MIA threat in both learning paradigms.
    Furthermore, we suggest a new taxonomy for MIAs based on the type of adversary access instead of the knowledge.
    \item We demonstrate limitations on the most used MIAs approaches through an empirical evaluation tested in tabular data for simple learning tasks. Most of the MIAs proposals focus on image data and complex tabular data which requires complex learning models (e.g.~\cite{Yeom, Passive-Active-WB-MIA,Enhanced-MIA-Game,rmia,MIA-first-prin}). 
    Although some prior studies have performed similar comparisons~\cite{MIA-in-MLaaS,central-fl-comp}, we expand our experiments by evaluating multiple MIAs approaches, incorporating defense mechanisms, and experimenting across a diverse range of tabular datasets. Furthermore, in FL scenario, we perform an outsider attack. In FL, if the adversary acts as the central aggregator, they hold a stronger position, as they can control the aggregation process by observing the learning dynamics over multiple rounds and extract more detailed information from participants' updates. In contrast, if the adversary is an outsider, they only have access to the final model. This scenario has been disregarded; as such, we adopt an outsider attack.  

    \item We evaluate the vulnerability of single-outs to MIAs. Singling-out an individual means that was uniquely identified by an adversary~\cite{k-anonymity}. Such data points present distinctive patterns, which may be prone to memorization. Thereby, the adversary's confidence is amplified and consequently, the risk of successful inference is increased.
    To our best knowledge, this has not been explored.

    \item We study the transferability of the attack models~\cite{Salem-ML-Leaks,MIA-in-MLaaS}. While most of literature assumes that the adversary possess detailed knowledge of the target model, it is important to understand to which extent attacks remain successful with less knowledge. Relying on different surrogate models allows for assessing the risk of MIAs in more realistic black-box scenarios.
    
\end{enumerate}


Our results show that MIAs on tabular data remain challenging: even overfitted models are only marginally more vulnerable than random guessing, while well-generalized models can be to some extent vulnerable. Additionally, model-free attacks fail to succeed against well-generalized models.
In FL, limited knowledge can be sufficient to perform attacks.
Notably, single-out data points increase attack success, heightening the vulnerability of unique records.
Furthermore, attack models are largely transferable, i.e., using different surrogate models increases attack effectiveness, suggesting that target model knowledge is not essential. 
In addition to the in-depth discussion of the caveats of each approach, we also provide potential future avenues in light of our findings.

To foster further research on this topic and facilitate reproducibility, we release the code for our analysis available at {\textit{anonymous link}}.


\section{Preliminary}~\label{sec:foudations}
Consider a classification problem (supervised learning setting), where a model $f$ with parameters $\theta$ maps inputs $x \in \mathcal{X}$ to an $n$-class probability distribution, $f_{\theta} : X \rightarrow [0, 1]^n$. The probability of class $y$ is then described as $f(x)_y$, in which $y$ can assume different dimensions depending on the learning task. Let $D$ be a training dataset, sampled from some underlying distribution $\mathbb{D}$, where $N$ is the number of data instances. We denote $f_\theta \leftarrow \mathcal{T}(D)$ as a $f$ trained model with parameters $\theta$ that is learned by running the training algorithm $\mathcal{T}$ on the training set $D$.

\textbf{Membership Inference Attacks.}
A successful MIA occurs when an adversary correctly determines if a record does or does not belong to the training set of the $f_{\theta}$ (target model). In the former case, the record is said to be a member, and in the latter case the record is a non-member. 
We define membership inference inspired in a standard security game~\cite{Yeom,jayaraman2020revisiting,MIA-first-prin,rmia}.

\begin{definition}[Membership Inference Game]
The game proceeds between a challenger and an adversary. Let $\mathbb{D}$ be  the data distribution and $\mathcal{T}$ the training algorithm.
\begin{enumerate}
    \item The challenger samples a training dataset $S \leftarrow \mathbb{D}$ and trains a model $f_\theta \leftarrow \mathcal{T}(S)$ on $S$. 
    \item The challenger flips a bit $b$. if $b=1$, it randomly samples a data point from the training set $(x,y) \leftarrow S$.
    Otherwise, it selects a point from the distribution $(x,y) \leftarrow \mathbb{D}$ such that $(x,y) \notin S$.
    \item The challenger sends the target model $f_\theta$ and target data point $(x,y)$ to the adversary.
    \item With access to the distribution $\mathbb{D}$, the adversary computes $Score_{MIA}((x,y);f_\theta)$, and outputs a membership prediction bit $\hat{b} \leftarrow MIA((x,y);f_\theta)$.
    \item The output of the game is $1$ if $\hat{b} = b$, and $0$ otherwise.
\end{enumerate}
\end{definition}

The prior knowledge of the adversary with respect to the target model, as well as the attack approach, are important factors that can greatly influence the success of MIAs.
By addressing the first factor, the adversary can have black-box, gray-box, or white-box knowledge. 

Black-box attacks refer to the case where an adversary has no access to the trained learning model. The literature of black-box MIA categorizes the adversary's knowledge into three levels. The minimal knowledge only includes the predicted label; in the next level, the adversary knows the probability vector of the target model; finally, the adversary can have additional access to the architecture and data distribution of the target model. This last assumption can also be considered
a grey-box attack as additional knowledge is assumed.


In a white-box attack, adversaries have complete knowledge about a target model (architecture and parameters), such as gradients and outputs of the hidden layers, which facilitates the membership inference. Although white-box attacks are more common, black-box attacks are more realistic for ML systems as essential model information, such as model implementation details, is often confidential and protected from normal users interacting with the system. 
Henceforth, we focus on the two primary threat scenarios: white-box and black-box attacks.

\textbf{Centralized vs Federated Learning.}
The most fundamental form of MIA was initially applied in centralized learning~\cite{Shokri-MIA-Against-ML}. In other words, the target model ($f_{\theta}$) is trained locally on a single machine using the full training dataset ($D$). 
FL~\cite{mcmahan2017communication} allows multiple participants to collectively train the same model without directly sharing their private data. This mechanism includes the participants involved in the model training and a central aggregator who collects the parameters received from the participants after each (or multiple) training round(s) and updates the global model. 
Formally, $f_{\theta}$ is the global model held by the central aggregator, and each participant, $j \in \{1,2,...,C\}$ holds a local instance of the model parametrized by $\theta_j$ and a private dataset $D_j=\{(x_{j,i},y_{j,i})\}^{n_j}_{i=1}$ sampled from the local distribution $\mathbb{D}_j$.


Regarding data distribution among participants, the common learning paradigm in FL includes Horizontal (HFL) and Vertical Federated Learning (VFL). In the former, each participant keeps the same features, but the sample records differ, $D_i \cap D_j=\varnothing,\forall_{i,j}$. In the VFL setting, participants have different features for the same sample records, $ID(D_i) = ID(D_j)$, where 
$ID(\cdot)$ represents the set of shared entity identifiers (e.g., user IDs). 


In both centralized and federated scenarios, once the training process is completed, $f_{\theta}$ outputs for each input $x$, a probability vector $p \in \mathbb{R}^k$ of the form $p = (p_1,p_2,...,p_k)$, where $p_i \in [0,1] \forall i$ and \(\sum_{i=1}^{k} p_{i}=1\). The prediction class label for a feature vector $x$ is the class with the highest probability value in $p$, thus \(\hat{y}=argmax_{i \in \{1,...,k\}}f_\theta(x)_i\).
The output distribution $p = f_\theta$ is often informative about whether a point ($x,y$) was included in the training dataset $D$.
During the attack, a membership score $Score_{MIA}((x,y);f_\theta)$ is assigned to each pair of $((x,y);f_\theta)$. Then, a membership  bit is outputted by comparing the score with a selected threshold $\beta$ which is used to distinguish members from non-members.

\begin{equation}~\label{eq:mia_attack}
\text{MIA}(x; \theta) = \mathbbm{1}\!\left[\text{Score}_{\text{MIA}}(x; \theta) \geq \beta \right],
\end{equation}

where \(\mathbbm{1}(\cdot)\) is an indicator function that results in 1 if the event occurs, otherwise 0.

\textbf{MIAs in overfitted vs well-generalized target models.}
A successful MIA is typically associated with a high level of overfitting of the target model~\cite{SoK,Salem-ML-Leaks,Shokri-MIA-Against-ML}. Overfitting occurs when a model fits too closely to its training data, resulting in high train accuracy results but fails to make accurate predictions in unseen data. This phenomenon is also associated to memorization, in which larger models are prone to memorizing the training data, thus making them more vulnerable to MIAs~\cite{MIA-first-prin, li2024privacy}. Consider a simple loss function $L(f_\theta(x)_y)$, where $R(\theta)=1/n\sum^{n}_{i=1}L(f_\theta(x)_y)$ is the average loss on the training data and $\hat R(\theta)$ the expected loss on unseen data. If the margin of error between $\hat R(\theta)$ and $R(\theta)$ is large and positive, it is said that the model is overfitting. Nevertheless, it has been shown that although overfitting is sufficient it is not necessary for the success of such attacks~\cite{Yeom,sablayrolles2019white}; well-generalized models can leak membership information, when vulnerable data points that cause unique influences on the learning target model, such as outliers, are present~\cite{Long-MIA-Well-Gen}.

\section{Literature Review}~\label{sec:sota}
This section provides a literature overview of MIAs considering two learning paradigms. First, we present the scenario in which the target model is trained centrally, including different access types, 
as well as proposed mitigation strategies. Second, we examine MIAs and respective defenses in the FL environment. 

\subsection{MIAs in Centralized Learning}~\label{subsec:CentralLearning}
\begin{figure}[!ht]
\centerline{\includegraphics[width=\columnwidth]{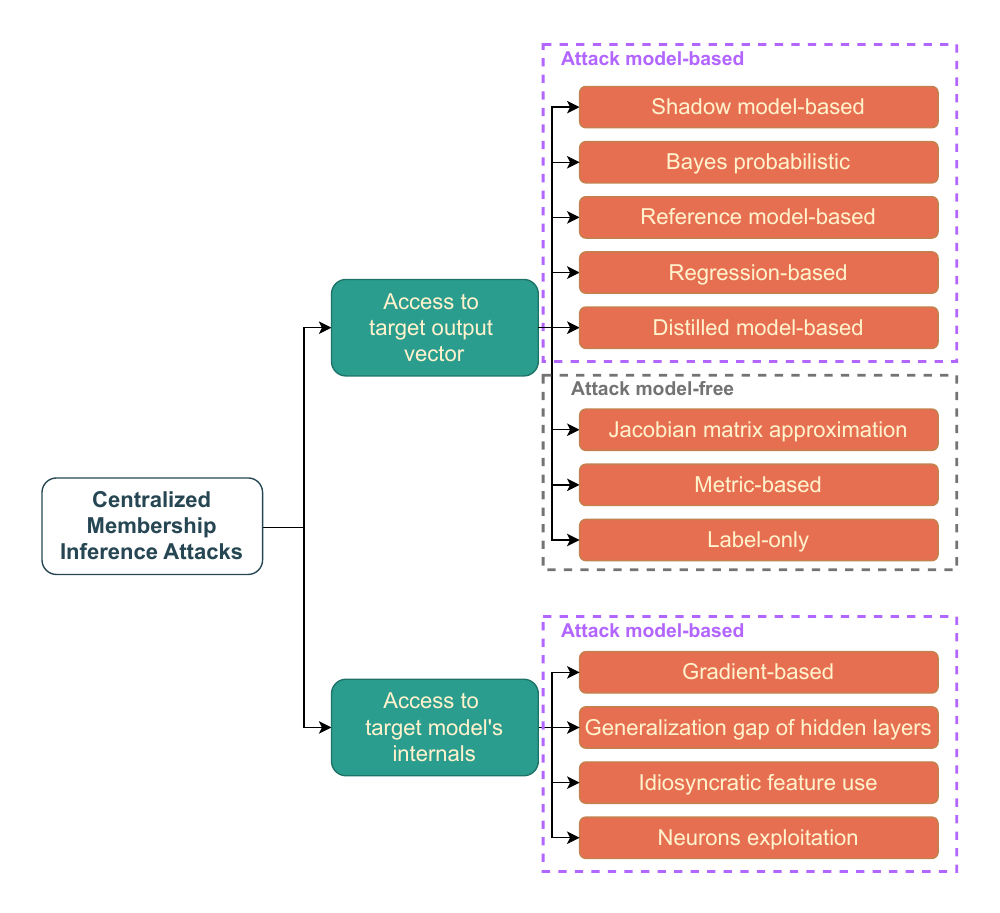}}
\caption{Taxonomy for Membership Inference Attacks conducted in centralized learning paradigm.}
\label{fig:mia_taxonomy_tree}
\end{figure}

MIAs are commonly divided with respect to the type of knowledge of the adversary, that is, black-box or white-box~\cite{SoK,rao2024privacy}.
However, a white-box adversary can perform any black-box attack. Therefore, we propose a new categorization based on the type of information available to the adversary. Figure \ref{fig:mia_taxonomy_tree} presents the new taxonomy of MIAs in a centralized setting. We divide it into two distinct target model access scenarios. \textit{Access to target output vector}, in which an adversary only has access to predicted labels or probabilities, and \textit{access to target model's internals}, in which an adversary has access to model architecture. 

The proposed division provides a more precise categorization. 
By dividing attacks according to the minimal access they require, the taxonomy avoids the overlap between white- and black-box knowledge. 

\subsubsection{Access to target output vector}
There are three main types of output vectors: probability, logit and label. When the first two are available, the attack is usually performed by training surrogate models. Thus, we divide the attacks into model-based and attack model-free strategies.

\textbf{Attack model-based.} 
One of the pioneers MIAs in ML~\cite{Shokri-MIA-Against-ML} perform the attack by training multiple shadow models that mimic the behavior of the target model. Under a black-box setting with access to output probabilities and partial architectural assumptions, multiple shadow models are trained on datasets sampled from the same distribution as the target model. As it is known which members (training records) and non-members (holdout records), then an attack training set is built consisting of rows of the form $\{\textit{shadow model's top probabilities, ground-truth label, } x\}$ (or predicted label) for each shadow model record, where $x$ is 1 if the record is a member of the shadow dataset, and 0 otherwise. Thus, this attack is typically converted into a classification problem in which the final attack model is used to perform the MIA. Multiple studies have demonstrated its effectiveness~\cite{MIA-in-MLaaS, li2021membership, niu2025comparing}.

However, this approach presents two main challenges. First, the success of the attack depends on the knowledge of the data distribution. 
Second, it requires multiple shadow models, which is computationally expensive, especially when the model under attack is large. Therefore, several researchers show improvements in MIAs by restricting knowledge assumptions and using limited shadow models -- in some cases, just one~\cite{Yeom, Salem-ML-Leaks}. In addition, it has been shown that MIAs can be performed through a data transferring attack~\cite{Salem-ML-Leaks} or model transferability~\cite{MIA-in-MLaaS}.

Given all the necessary assumptions of the former shadow-based attacks, a different strategy shows that black-box attacks perform as well as white-box
attacks in an optimal asymptotic setting~\cite{sablayrolles2019white}. This strategy depends only on the loss of the classifier and uses Bayes optimal approximations to derive concrete MIAs. Given a parameter $\theta$ and a sample $z$, the optimal membership is inferred by computing the optimal decision rule that maximises the probability of correct membership classification. 
We classify it as Bayes probabilistic attack.

Nevertheless, one pitfall is that it only measures the means of the distributions, which means that the attacks can be sub-optimal if different samples’ loss distributions have very different scales and spreads~\cite{MIA-first-prin}. A more efficient parametric approach is then proposed with a principled and well-calibrated Gaussian likelihood estimate.
The attack is performed as follows. First, the adversary trains many reference models, half of which contain the target example $(x,y)$ (IN models) and half will not (OUT models). Next, the adversary queries each reference model $f_\theta'$, to compute the logits corresponding to the correct class, $f_\theta'(x)_y$, and
fits a Gaussian $\mathcal{N}(\mu_{IN},\sigma^2_{IN})$ to the logits for all models containing the example (and similarly for the OUT models). The adversary then computes the probability density function of each Gaussian and the likelihood ratio, which serves as the membership score. 

Here, reference models are identical to shadow models. The difference lies in their purposes: the goal of shadow-based attacks is to mimic the behavior of the target model, while the objective of reference-based attacks is to analyse the behavior of the IN and OUT models to estimate the likelihood of the output of the target model under the IN/OUT hypotheses. 
Some studies have then been published on this~\cite{Hierarchical-tree-based, yichuan2024assessing, suri2024parameters,tao2025range,rmia} demonstrating higher performance than shadow-based attacks.
However, the perfect attack should remove the randomness in the training data. In this setting (leave-one-out attack), the adversary trains reference models $f_\theta'$ on $D\setminus\{x,y\}$~\cite{Enhanced-MIA-Game, suri2024parameters}. This requires assuming that the adversary
already knows exactly the $n-1$ data records in $D\setminus\{x,y\}$, which is a strong assumption, yet it is relevant for privacy auditing.

One limitation is that a simple modification to the score, called difficulty calibration, can drastically improve the reliability of the attack~\cite{watson2021importance,he2024difficulty}. A low loss sample is not necessarily a member; it could also be due to its low difficulty. Typically, difficulty calibration assumes that outputs from reference models can effectively represent the difficulty of target samples, thus improving existing
MIAs.
Another limitation is that reference-based attacks require the training of multiple reference models, which is particularly critical when the target model is large. Thus, an agnostic approach was proposed to overcome such a problem~\cite{bertran2023scalable} which does not involve the training of shadow or reference models, only the access to the member and non-member set to perform the attack by training a quantile regression model.

A different approach uses the advantage of knowledge distillation to perform MIAs in which a "teacher" model transfer the knowledge to a "student" model. 
Reference-based attacks have been adapted to knowledge distillation~\cite{Enhanced-MIA-Game,galichin2025glira,jagielski2023students}. In particular, self-distillation is used, which means that the teacher and student datasets are identical. 
MIAs have been shown to be stronger when student and teacher sets are similar~\cite{jagielski2023students}.

\textbf{Attack model-free.}
A promising direction involves using the Jacobian matrix of the training samples. The idea is to quantify the prediction sensitivity with the Jacobian matrix which reflects the relationship between each feature's perturbation and the corresponding prediction's change~\cite{liu2022membership,central-fl-comp}. Training samples typically have lower prediction sensitivities than testing data. First, the Jacobian matrix is approximated and the prediction sensitivity with respect to the target model is derived. Then, the target samples are clustered according to sensitivity, and the cluster with the lowest mean sensitivity is identified as the members. Thus, an unsupervised attack is applied. Unlike shadow attacks, in which members and non-members are known, unsupervised attacks follow the strategy of an adversary who does not know whether a data sample belongs to the training data. However, this method requires finding the best $\epsilon$ to derive an approximation of the Jacobian matrix, which may be time-consuming.

By comparison, metric-based approaches allow MIA to be performed without any assumptions~\cite{MIA-ML-Survey}. 
For example, MIA based on the prediction correctness metric requires only knowledge of the predicted labels. The attack follows this principle: if the target model correctly predicts a record, then the adversary infers it as a member, otherwise as a non-member~\cite{Yeom}. 
An adversary can also infer an input record as a member if its prediction loss is smaller than the average loss of all training members. 
This requires computing the cross-entropy loss of the outputs of the target model and comparing it with a selected threshold to distinguish members from non-members~\cite{Yeom}. 

Different approaches include inferring a record as a member if its maximum probability is higher than a threshold or its prediction entropy is lower than a threshold~\cite{Salem-ML-Leaks}. In the former case, the target model is trained by minimizing prediction loss over its training data, leading to a high confidence score of a training member’s prediction, while in latter case the prediction entropy distributions between training and test data are very different. The  target model’s outputs are generally less confident and more uncertain, resulting in higher entropy values compared to those for training samples.

However, prediction entropy does not consider any information about the ground-truth label, leading to a misclassification of members and non-members. For instance, a wrong classification with a probability score of one results in entropy equal to zero for an input record, and thus the adversary classifies it as a member. Yet, a record with a totally wrong classification is likely to be a non-member. Thus, the prediction modified entropy metric leverages the information of the ground-truth label~\cite{Systematic-eval}.


A big concern of the majority MIAs is that the confidence (or loss) scores of the model are exposed, which rely on continuous-valued predictions. For a higher level of protection, prior work has considered obfuscating the model's output by returning only the top label or modifying predicted values for a higher level of protection. Given a target model, the adversary queries an input record and obtains only the predicted label.
Despite this limited information, label-only attacks can still identify whether a record was in the original training data~\cite{Label-MIA,li2021label,he2025towards}.
Membership can be inferred by exploiting data augmentation and decision boundary distance to evaluate robustness of the target records to these input perturbations~\cite{Label-MIA,li2021label}.
Label-only attacks also benefit from difficulty calibration~\cite{watson2021importance}, making them more effective.

\subsubsection{Access to target model's internals}
Although previous attacks are carried out with limited knowledge, several researchers have shown how to improve the success of attacks in deep learning models~\cite{Passive-Active-WB-MIA,cohen2024membership,White-Box-MIA-DL,Stolen-Memories}. It is assumed that an adversary might have access to the internals of the target model, which typically constitutes a white-box scenario.

One of the first proposals demonstrate how to exploit the privacy vulnerabilities of the stochastic gradient descent (SGD) algorithm~\cite{Passive-Active-WB-MIA}. During training of the target model, SGD minimizes the loss function by updating parameters in small subsets of the training set. The intuition is to capture the distinguishability of each training data sample in the gradients of the loss function. The gradients tend to reveal more information about the membership than the layer outputs, even for well-generalised target models. 
Moreover, the gradients can be used to infer the influence values, i.e., the influence a data point in the training set has in the target model's prediction for a given test sample~\cite{cohen2024membership}. Larger gradients are correlated to larger influence, and therefore a high membership score. However, self-influence values require calculating the Hessian vector product, which is computationally costly. 

When the internals of the model are accessible, it is also possible to perform MIAs by evaluating the internal generalization gap of the hidden layer space~\cite{White-Box-MIA-DL}. 
This gap is estimated by calculating the KL divergence of the features in the hidden layers of the model in the training and test sets. Initial layers tend to increase the gap sharply, but it gradually decreases across the remaining layers.

Additionally, some records in the target model's training set may contain specific features that identify them uniquely and thus make them more susceptible to MIAs. In other words, a model's idiosyncratic use of features can provide
evidence for membership~\cite{Stolen-Memories}.
For instance, if a target model is trained to recognize a particular individual among multiple faces and only one contains a pink background, then this feature could serve as evidence of membership. This demonstrates that privacy issues resulting from overfitting do not necessarily manifest in the output behavior, but rather in the way the model uses features. A Bayes-optimal threat model is used to derive a set of parameters that profile idiosyncratic feature use, which are then used to construct a logistic attack model. 

Lastly, selecting the most influential hidden activations rather than all of them may improve the efficiency of MIA. This is because only a subset of neurons substantially impacts the target classification task~\cite{WB-Neurons}. 
KL divergence and random forest can be used to select the most influential hidden activations across all layers. Then, explainable AI techniques (e.g. SHAP) can be used to quantify the features' significance in each neuron and explain their directional influence. The disparity between the two neuron sets allows inferring the membership score.


\subsection{Defenses to MIAs in Centralized Learning}~\label{subsec:CentralDefenses}
Given the wide range of MIAs proposed in centralized learning, there is a growing need to develop mitigation mechanisms that effectively reduce their success.
Defenses against MIAs can be applied during the training phase or during the prediction phase~\cite{MIA-in-MLaaS}. They are typically classified as either model hardening or API hardening methods. The former aims to apply techniques that reduce the target model's level of overfitting. The latter aims to limit the information accessible in the prediction vector. 

\subsubsection{Model Hardening Defenses} This category covers regularization techniques, 
hyper-parameter tuning, differential privacy, model stacking, knowledge distillation, outlier detection, and lastly, k-anonymity.

\textbf{Regularization techniques} help to smooth model confidence distributions which may weaken MIAs. For instance, dropout reduces overfitting by reducing variance~\cite{Shokri-MIA-Against-ML,Salem-ML-Leaks,kaya2020effectiveness}. It randomly sets some nodes to zero during the training process. This effectively creates smaller networks that each need to learn to solve the network's task. Although this mitigation is generally efficient, it is limited to neural networks.
$l$2-regularization~\cite{Shokri-MIA-Against-ML,MIA-in-MLaaS,Label-MIA,li2021label} penalizes large weights for a better generalization and thus reduces the membership/non-membership confidence gaps. However, this must be carefully implemented to avoid damaging the performance of the model. 


Adversarial regularization aims to reduce the prediction loss. A common approach is Min-Max privacy game in which the target model is trained using an adversarial process~\cite{Adv-Regul,li2021membership}. Specifically, the gain of the MIA is used as a \textit{regularizer} for the target model classifier that controls the privacy-utility trade-off. 

In addition, adversarial training has been used in the context of MIAs defenses. Both adversarial training~\cite{MemGuard,li2024privacy} and data augmentation~\cite{cohen2024membership, Stolen-Memories,li2024privacy, li2021membership} involve adding certain examples to the training set to improve performance. However, applying adversarial training will make the model memorize more training samples, thereby causing more privacy leakages while data augmentation may be not effective against MIAs~\cite{li2024privacy}.

Multi-sample convex combination~\cite{White-Box-MIA-DL} leverages the generalization gap of the internal layers of the target model. This defense mechanism creates synthetic feature representations in hidden layers by linearly mixing the outputs of hidden units from different data points, resulting in new blended feature representations. This discourages the model from having highly specific memorized patterns in the hidden layers, thus obscuring whether a specific data point was part of the training set or not. This was designed to protect models in a white-box scenario.

Label smoothing or early stopping can also act as regularization techniques~\cite{kaya2020effectiveness}. However, label smoothing can inadvertently increase vulnerability to MIAs. By reducing prediction confidence, it can obscure overfitting signals yet amplify membership cues. Conversely, even though early stopping is not the most effective method, it is a consistent way of preventing MIAs: the earlier training ends, the less successful MIAs are.

\textbf{Hyper-parameter tuning} of ML models must be carefully defined. Certain configurations can lead to a higher risk of MIAs. 
Some models, namely tree-based, have already been tested with multiple combinations of hyper-parameters~\cite{Hierarchical-tree-based} to guide in the parameters selection.
In addition, attention should be given to the choice of target model. For example, the Naive Bayes classifier has been shown to be more resilient to these attacks as it usually does not overfit~\cite{MIA-in-MLaaS}.

\textbf{Differential privacy} is a common mitigation technique~\cite{Label-MIA,Stolen-Memories,MIA-first-prin,watson2021importance,he2024difficulty,li2021label,cohen2024membership}; however, several studies have highlighted the difficulty of achieving a satisfactory privacy-utility trade-off. In fact, the privacy budget $\epsilon$, which represents the degree of privacy protection, 
when small, it decreases the effectiveness of the adversary, but also reduces the utility of the target model. Some variations have been proposed to obtain a satisfactory privacy-utility trade-off. For instance, transfer learning in which the knowledge learned from a task is reused can be combined with differential privacy~\cite{Label-MIA}. Also, Differentially Private SDG, adds calibrated noise to gradients. Differential privacy can be applied only on the top layers~\cite{Stolen-Memories}. Furthermore, regularization is also used for objective perturbation in differentially private ML.



\textbf{Model stacking} is based on ensemble learning, which combines multiple ML models (base models) to construct a final model~\cite{Salem-ML-Leaks,li2021membership}. The base models obtain the outputs and concatenate them to apply the result to the final model, which predicts the final output. All base models are trained on disjoint sets of data. However, this strategy is easily circumvented in a white-box setting as the base models are available to the adversary~\cite{Stolen-Memories}.

\textbf{Knowledge distillation} is commonly used for model compression but it also protects against
membership attacks by thwarting the access of the resulting models to the private training data~\cite{shejwalkar2021membership,jagielski2023students,tang2022mitigating,kaya2020effectiveness}. The intuition is that since the student model is not directly optimized over the private set, their membership may be protected. However, this method can only be applied when a public dataset is available for training.

\textbf{Outlier identification} estimates whether a given instance is an outlier. An outlier can be more likely to be a vulnerable target record when it is in the target model's training set, even if the model is already well-generalized~\cite{Long-MIA-Well-Gen}. After identifying the outliers, the data points that do not significantly contribute to the utility of the model are removed.

\textbf{$k$-anonymity} allows the training dataset of the target model to be made $k$-anonymous prior to training so that the impact of a single instance is hidden among $k$ other records~\cite{MIA-in-MLaaS}. 
Similar to outliers identification, the unique signature instances may be more susceptible to MIAs.


\subsubsection{API Hardening Defenses} This category aims to modify the model's external outputs, which includes restriction of prediction vector information and adding noise to the prediction vector.

\textbf{Restriction of prediction vector information} can be achieved using several strategies, for example by: restricting the prediction vector to the top $k$ classes by adding a filter to the last layer of the target model~\cite{MIA-in-MLaaS}; decreasing the precision of the prediction vector by rounding the classification probabilities down to $d$ floating-point digits; and, increasing the entropy of the prediction vector by adding or modifying the softmax layer and increasing its normalizing temperature to above zero~\cite{Shokri-MIA-Against-ML}. 


\textbf{Adding noise to the prediction vector} means that the crafted noise is carefully added to each confidence score vector of the target classifier~\cite{MIA-in-MLaaS,MemGuard}.
Since the adversary may know the defense mechanism, but the defender does not know the attack classifier, it is essential that the noise is added randomly; otherwise, the adversary could learn it and adapt their MIA technique. 
However, the distance between the noisy and true prediction vectors should remain minimal to preserve the optimal trade-off.
Although stated as effective in black-box scenarios, this defense does not modify the generalization gap in the hidden layers of the target model, thus does not protect against adversaries that have access to the model's internals (white-box)~\cite{White-Box-MIA-DL}.

Despite the notable differences in defensive approaches between model and API hardening, some can be combined to enhance output protection. For example, generalization and perturbation can be applied together~\cite{Long-MIA-Well-Gen}.


\subsection{MIAs in Federated Learning}

FL was initially proposed to address data privacy concerns, such as maintaining the confidentiality of sensitive information~\cite{fl_google}. However, multiple studies have shown its vulnerability to MIAs~\cite{MIA-in-MLaaS, Passive-Active-WB-MIA,shadow-update-based-fl,MIA-BAD-FL,CS-MIA,efficient-trend-based-fl,MIA-FL-Bias,source-inf-fl,nikolaidis2025study,central-fl-comp,zhao2021user}.

MIAs in FL setting differ from the centralized case mainly in the adversary's level of participation and the attack phase.
In centralized learning, the adversary is typically passive, having only access to the final model. 
In contrast, in FL, the adversary can be one of the participants involved in the local or in the global model training. 

A local adversary owns its private data and can access global models during rounds. The goal is to determine a record's membership of all participants' training data. If the adversary acts as the central aggregator, they have an advantage as they can control the aggregation process and thus extract more information. The adversary can carry out passive or active attacks. In the former, the adversary only observes the information exchanged between the participants and the central aggregator. In an active attack, the adversary directly interferes by altering the global model or aggregates specific participants each round.

Furthermore, there is a distinction between an insider and an outsider adversary. While an insider adversary can be described as someone with knowledge of the FL internals (local and global adversary), an outsider adversary is someone with access only to the global model~\cite{MIA-in-MLaaS}.

Additionally, when an adversary is involved in the FL training process, they have access to the exchanged gradients, intermediate computations, and historical versions of the target model. However, a simple MIA in an FL environment does not enable an adversary to identify the specific participant to which the target record belongs. 
If the adversary can predict the target participant, this is known as a Source Inference Attack~\cite{source-inf-fl}. 

\begin{figure}[ht!]
    \centering
    \centerline{\includegraphics[width=\columnwidth]{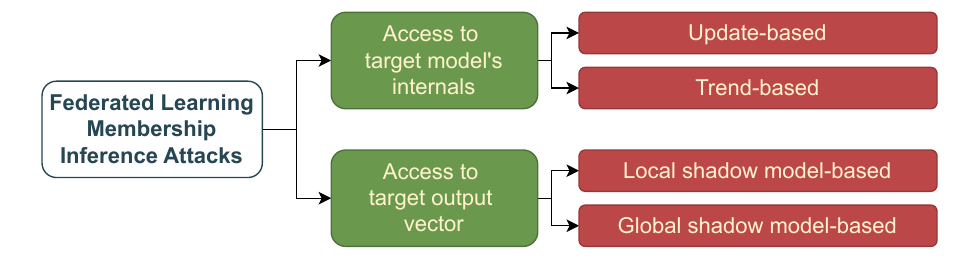}}
    \caption{Taxonomy for Membership Inference Attacks conducted in federated learning paradigm.}
    \label{fig:mia_taxonomyfl}
\end{figure}

We focus on MIAs in HFL given the immature state of VFL in this context. 
Figure~\ref{fig:mia_taxonomyfl} presents a new taxonomy for FL setting. We propose a new category given the type of information an adversary has access to. Hence, we divide into access to the target model's internals and target output vector. 
In the former, we include the well-known update-based and trend-based approaches. In the latter, we propose local and global shadow model categories.  

\subsubsection{Update-based attacks}
In these attacks, the adversary exploits the information exchanged between the participant and the server, such as the model gradients and weights. Common approaches include exploitation of the SGD algorithm, shadow models, and data augmentation.

As previously mentioned in the centralized paradigm, leveraging the SDG algorithm can lead to MIAs. This has also been applied in the context of a collaborative environment~\cite{Passive-Active-WB-MIA}.
In an active attack, the aim is to forcefully decrease the gradient of loss in the training data. The adversary runs the attack on the target record and updates the local model parameters to increase the loss at this data point. This can be achieved by simply adding the gradient to the parameters. The adversary then uploads the adversarially computed parameters to the central server, which aggregates them with the parameter updates of the other participants. The inference model can detect when a participant's local SGD abruptly reduces the loss gradient on a record. The accuracy of the attack tends to be higher if the adversary acts as the central aggregator rather than as a participant, due to the observation of aggregated parameters from multiple participants.

Whereas this approach updates the parameters in the direction of gradient descent, gradient ascent aims to update the parameters in the opposite direction. The purpose of gradient ascent is to induce the target participant to decrease the loss of its local model at the target data points through SGD, as the adversary maliciously increases the loss by using gradient ascent~\cite{wang2023gbmia}. Besides, this new proposal works is attack model-free as the MIA is determined based on the gradient norm comparison. 

Another transposed approach from centralized to the distributed paradigm is shadow modelling. In this context, it can be applied to the local models. MIAs via shadow models have been applied in sequential FL~\cite{shadow-update-based-fl}, that is, each participant trains the local model and sends the updated parameters sequentially, not in parallel.
This cyclic nature is more vulnerable to these attacks, as the adversaries can observe the dynamics of the learning process. Here, the adversary is a passive participant who has access to the architecture and data distribution of the target model. Unlike most MIAs in FL, the attack is performed during inference time. 

A sophisticated approach, based on data augmentation~\cite{zhao2021user}, aims to infer which participant a data point belongs to. High-quality fake samples are generated using a local generative adversarial network (GAN). The parameters of the local model are copied to the GAN which are updated synchronously. The objective of the data augmentation phase is to complete the training set for the attack model. This approach has been shown to be efficient when a wireless network monitor is available.

\subsubsection{Trend-based attacks}
In such attacks, the adversary exploits the trajectory of model output by analyzing the evolving patterns of the learning process. 
Different trends (behaviors) can be used to distinguish members from non-members. We identify three main trends: confidence, prediction correctness and bias. 


An adversary who observes the confidence values of the probability vector of the target model can build a temporal series of the fluctuations of these values among various rounds~\cite{CS-MIA}. These patterns reveal membership information, as the confidence of member samples tends to rise faster than that of non-members. To perform the attack, the adversary first has to generate shadow confidence series on a shadow dataset for members and non-members, and then use the results to build the attack model. 
The confidence of the model is captured using a modified prediction entropy. 

A variation of the prediction correctness can also be used in this context to infer membership~\cite{efficient-trend-based-fl}. When collecting the target models, a score can be assigned to the correct label for each input. Then, the adversary analyses the temporal evolution of the true label scores between members and non-members to infer membership. This strategy requires less computation power and memory.

The weight parameter, exchanged between the participants and the central aggregator, is frequently exploited in MIAs in FL, but it leads to significant overhead~\cite{MIA-FL-Bias}. To overcome this, the bias parameter can be analyzed over various training rounds. While the weight is defined by the direction of the decision surface, the bias determines the distance between the probability vector and the ground-truth label. This bias-based attack can operate at the local (participant) and global (aggregator) levels. A classification model with bias values is used to infer the MIA in the first case, while in the latter, the adversary compares bias differences for each local model and then counts the greater values that occur in each participant. 

\subsubsection{Local shadow model-based}
Shadow approach can also be leveraged in FL as in centralized. In this setting, an adversary can query the local learning model to get a prediction on a given sample, but has no other access to the model and its weights and gradients. In addition, the adversary cannot get any knowledge about the global training process. A new shadow model-based attack was proposed using a batch-wise strategy~\cite{MIA-BAD-FL}. For every shadow model, evaluate the entire shadow dataset batch-wise, and record the batch-wise loss with the seen/unseen label to build the attack dataset. A simple binary classifier is then trained on the attack dataset as the attack model.

\subsubsection{Global shadow model-based} 
In FL, it is generally assumed that the adversary is either one of the participants or the central aggregator. Accordingly, attacks typically rely on access to a model's internal parameters.
Nevertheless, an FL model exposed via an API can still be vulnerable to MIAs. Such attacks can be carried out in a black-box setting, using only the global model outputs~\cite{MIA-in-MLaaS}. 
Thus, given access to the probability vectors, the adversary can identify whether a training instance belongs to the FL system (outsider attack) or identify target records from other participants (insider attack).

\subsection{Defenses to MIAs in Federated Learning}~\label{subsec:FLDefenses}
Multiple defenses originally proposed against MIAs in centralized learning can be adapted for FL. 
Namely, regularization techniques, differential privacy and knowledge distillation. Other defense strategies have been developed specifically focusing on the nature of FL. This includes secure aggregation and digestive training.

\textbf{Regularization techniques} such as dropout~\cite{CS-MIA,central-fl-comp}, $L1$ and $L2$-regularization~\cite{MIA-FL-Bias}
have been directly applied in the FL paradigm. On the other hand, adversarial regularization has been adapted. For example, a small simulated adversary network can be embedded as a regularization punishment~\cite{su2021federated}. However, this approach requires two networks to be trained alternately with the gradient modification algorithm, which may be more resource-expensive.

\textbf{Differential privacy}~\cite{MIA-FL-Bias, CS-MIA, source-inf-fl, shadow-update-based-fl} and \textbf{knowledge distillation}~\cite{central-fl-comp,CS-MIA} can also be implement as in centralized learning. Nevertheless, although knowledge distillation has been shown to be effective against MIAs with a minor performance impact on the target model, it requires participants to spend nearly twice as long to locally train teacher models and the student model in turns, leading to a significant performance overhead to resource-constrained FL device~\cite{CS-MIA}.

\textbf{Secure aggregation} aims to ensure that the server can successfully aggregate the global model without accessing participants' model updates. Each participant must mask their local model with pairwise random keys before sending updates to the server. This pairwise random masking is only removed when the server aggregates multiple participants~\cite{MIA-FL-Bias,CS-MIA}. In addition, homomorphic encryption can be leveraged to mask participants' update parameters, in which a parameter selection method is added to the aggregator to choose participants' updates with a certain probability to avoid higher performance impacts~\cite{bai2021method}.
However, these methods typically involve high communication and computation cost.


\textbf{Digestive neural network} aims to defend against MIAs by transforming input and skewing updates~\cite{lee2021digestive}. The private data owned by each participant will pass through this network and then train the FL. A mini-batch is transformed within the digestive neural network into another form. This digested mini-batch contains features useful for classification and the maximum distance from the original mini-batch. Thus, each participant collects the updates using the digested mini-batch.

Lastly, the following additional defenses are suggested: noise injection in the training data of each node to distort the resulting model; randomization of the order of the nodes in each cycle; and reduction of the number of training epochs at each node~\cite{shadow-update-based-fl}.

\subsection{Membership Inference Evaluation}~\label{sec:miaeval}
Regardless of the attack strategy, whether supervised (classification or regression~\cite{bertran2023scalable,galichin2025glira}) or unsupervised (clustering~\cite{liu2022membership,Passive-Active-WB-MIA}), the evaluation follows a classification problem, assessing how well the inferred groups distinguish between members and non-members. 
All attack evaluations are compared to a clueless adversary. This adversary makes random guesses about the membership and is expected to get an accuracy of 0.5.

MIAs were initially evaluated using \textit{accuracy} to capture overall correctness. Also, for a more comprehensive assessment, complementary measures have been used, such as \textit{precision}, which quantifies how many truly members belong to the training set, and \textit{recall} that measures the attack coverage, i.e. the amount of actual members that the adversary successfully detects~\cite{Shokri-MIA-Against-ML, Yeom, MIA-in-MLaaS, Stolen-Memories}. 
Other similar evaluation metrics include \textit{F1-score}~\cite{WB-Neurons,CS-MIA} and \textit{mean average precision} (mAP)~\cite{sablayrolles2019white}.

These evaluation metrics have been applied when the attack dataset is balanced, which means that the portions of members and non-members are similar.
\textit{Membership advantage}, obtained through the difference between \textit{true-positive rate (TPR)} and \textit{false-positive rate (FPR)}~\cite{Yeom} quantifies how much better an adversary is than random guessing at distinguishing members from non-members, in which a difference of zero corresponds to the random guess. Unlike precision/recall, the analysis of TPR/FPR is independent of the prevalence of members in the population, and thus may be more appropriate when this prevalence is unknown.

However, such metrics are average-case and assign equal cost to false positives and false negatives. Thus, \textit{Area Under the Curve (AUC)}~\cite{Salem-ML-Leaks} and \textit{Receiver Operating Characteristic (ROC)}~\cite{MIA-first-prin} were suggested to MIAs assessment since it emphasizes positive predictions over non-membership predictions. It compares the attack's TPR and FPR for all possible decision threshold. However, ROC curve should not be summarized through AUC, since AUC averages performance over all FPR, including high error rates that are irrelevant in practical attack scenarios. 
Hence, TPR@Low FPR was recommended to overcome this using for instance 0.001\% or 0.1\% as fixed FPR~\cite{MIA-first-prin}.

A concern remains with the previous MIAs evaluation metrics: they mostly focus on positive membership. But detecting non-membership is still relevant. In particular, FPR or \textit{False Alarm Rate (FAR)} shows how non-members are often misclassified as members~\cite{Rezaei2021}. 
It should be noted, however, that achieving reliable MIA with high accuracy and low FAR is difficult. Then a difficulty calibration was proposed to balance accuracy and FPR with the aim of achieving a better separation between member and non-member scores~\cite{watson2021importance}.
Similarly, rather than relying on a fixed confidence threshold of 50\%, a dynamic threshold is set based on a predefined false-positive tolerance \(\alpha\), which is set to the \(\alpha^{th}\)-percentile of confidence scores from a sample. This ensures a controlled FPR level with differing calibration or confidence distributions~\cite{Stolen-Memories}. 

\subsection{Summary}
Given the vast state-of-the-art in this topic, we provide in Table~\ref{tab:summary} the proposed MIAs for both learning paradigms. We specify whether it uses deep learning (DL) or traditional learning (TL), tabular (T) or/and image (I) data types, passive (P) or/and active (A) adversary strategy.

Active attacks in centralized learning can be leveraged through poisoning attacks~\cite{chen2022amplifying} 
or even through adversarial attacks~\cite{MemGuard,li2024privacy}, which typically enhances MIAs. However, such attacks are not within the scope of this paper. 

Note that despite some approaches not being named, we suggest names based on their characteristics. 

\begin{table*}[ht!]
\scriptsize
\resizebox{\textwidth}{!}{%
\begin{tabular}{llllllllll}
\toprule
\textbf{\begin{tabular}[c]{@{}l@{}}Learning\\ Paradigm\end{tabular}} & \textbf{Year} & \textbf{\begin{tabular}[c]{@{}l@{}}Approach \\ Name\end{tabular}} & \textbf{\begin{tabular}[c]{@{}l@{}}Attack\\ Classification\end{tabular}}   & \textbf{\begin{tabular}[c]{@{}l@{}}Learning\\ Type\end{tabular}} & \textbf{\begin{tabular}[c]{@{}l@{}}Adversary\\ Knowledge\end{tabular}} & \textbf{\begin{tabular}[c]{@{}l@{}}Adversary\\ Strategy\end{tabular}}  & \textbf{Defenses}         & \textbf{\begin{tabular}[c]{@{}l@{}}Data \\ Type\end{tabular}} & \textbf{Code}              \\ \hline
\multirow{23}{*}{Centralized}                                        & 2017          & MIA~\cite{Shokri-MIA-Against-ML}                                                               & Shadow model-based                                                         & DL                                                               & \CIRCLE                                                 & P                                                                       & \checkmark & T \& I                                                        & x \\
                                                                     & 2018          & LOSS~\cite{Yeom}                                                        & Shadow model-based                                                         & DL \& TL                                                         & \LEFTcircle                                             & P                                                                     & x                         & T \& I                                                        & x     \\
                                                                     & 2018          & ML-Leaks~\cite{Salem-ML-Leaks}                                                          & Shadow model-based                                                         & DL \& TL                                                         & \CIRCLE                                                 & P                                                                  & \checkmark & T \& I                                                        & \checkmark     \\
                                                                     & 2018          & GMIA~\cite{Long-MIA-Well-Gen}                                                              & Reference model-based                                                      & DL                                                               & \CIRCLE                                                 & P                                                & \checkmark & T \& I                                                        & x     \\
                                                                     & 2019          & Bayes Calibrated Loss~\cite{sablayrolles2019white}                        & Bayes Probabilistic                                    & DL \& TL                                                         & \LEFTcircle                                             & P                                   & x                         & I                                                             & x     \\
                                                                     & 2020          & Prediction uncertainty~\cite{Passive-Active-WB-MIA}                                            & Gradient-based                  & DL                                                               & \Circle                                                 &  P                                             & x                         & T \& I                                                        & x     \\
                                                                    
                                                                     & 2020          & Idiosyncratic MIA~\cite{Stolen-Memories}                                                 & Idiosyncratic feature use                                                  & DL                                                               & \Circle                                                 & P                                           & \checkmark & T \& I                                                        & x \\
                                                                     &  2021 & Privacy risk score~\cite{Systematic-eval}
                                                                     & Metric-based & DL & \CIRCLE & P & \checkmark & T & \checkmark \\
                                                                     & 2021          & Confidence-vector MIA~\cite{Label-MIA}                                             & Label only                                                                 & DL                                                               & \CIRCLE                                                 & P                                                     & \checkmark & T \& I                                                        & \checkmark     \\
                                                                     & 2021          & Boundary~\cite{li2021label}                                                          & Label only                                                                 & DL                                                               & \CIRCLE                                                 & P                                                     & \checkmark & I                                                             & x     \\
                                                                     & 2022          & Difficulty Calibrated Loss~\cite{watson2021importance}                                        & Reference model-based \& Label-only & DL                                                               & \Circle                                                 & P                                                                & \checkmark & T \& I                                                        & \checkmark     \\
                                                                     & 2022          & Attack R, Attack L~\cite{Enhanced-MIA-Game}                                                & Reference model-based                                                      & DL                                                               & \CIRCLE                                                 & P                                          & \checkmark & T \& I                                                        & \checkmark     \\
                                                                     & 2022          & Attack D~\cite{Enhanced-MIA-Game}                                                          & Distilled model-based                                                      & DL                                                               & \CIRCLE                                                            &                                              P                      & \checkmark & T \& I                                                        & \checkmark     \\
                                                                     & 2022          & ASTER~\cite{liu2022membership}                                                             & Jacobian matrix approximation                                              & DL \& TL                                                         & \LEFTcircle                                             & P                                                        & x                         & T \& I                                                        & x     \\
                                                                     & 2022          & SIF~\cite{cohen2024membership}                                                               & Gradient-based                                                             & DL                                                               & \Circle                                                 & P                                            & \checkmark & I                                                             & \checkmark     \\
                                                                     & 2023          & Quantile~\cite{bertran2023scalable}                                                          & Regression-based                                                           & DL                                                               & \CIRCLE                                                 & P                                                         & x                         & T \& I                                                        & \checkmark     \\
                                                                     & 2023          & GapMIA~\cite{White-Box-MIA-DL}                                                            & Generalization gap of hidden layers                                        & DL                                                               & \Circle                                                 & P                                                                 & \checkmark & T \& I                                                        & x     \\
                                                                     & 2024          & RMIA~\cite{rmia}                                                              & Reference model-based                                                      & DL \& TL                                                         & \CIRCLE                                                 & P                                        & x                         & T \& I                                                        & \checkmark     \\
                                                                     & 2024          & GLiRA~\cite{galichin2025glira}                                                             & Distilled model-based                                                      & DL                                                               & \CIRCLE                                                 & P                                                       & x                         & I                                                             & x \\
                                                                     & 2024          & LiRA-L~\cite{suri2024parameters}                                                            & Reference model-based                                                      & DL \& TL                                                         & \CIRCLE                                                 & P                                                     & x                         & T \& I                                                        & \checkmark     \\
                                                                     & 2024          & IHA~\cite{suri2024parameters}                                                               & Gradient-based                                                             & DL \& TL                                                         & \Circle                                                 & P                                              & x                         & T \& I                                                        & \checkmark     \\
                                                                     & 2024          & RAPID~\cite{he2024difficulty}                                                             & Reference \& shadow model-based                                            & DL                                                               & \CIRCLE                                                 & P                                                        & \checkmark & I                                                             & \checkmark     \\
                                                                     & 2024          & ShapMIA~\cite{WB-Neurons}                                                           & Neurons exploitation                                                       & DL                                                               & \Circle                                                 & P                               & x                         & I                                                             & x     \\
                                                                     & 2025          & RaMIA~\cite{tao2025range}                                                             & Reference model-based                                                      & DL                                                               & \CIRCLE                                                 & P                                                  & x                         & T \& I                                                        & \checkmark     \\ \hline \midrule
\multirow{8}{*}{Federated}                                           & 2020          & Sequential MIA~\cite{shadow-update-based-fl}                                                    & Update-based $\dagger$                                        & DL                                                               & \Circle                                                 & P                                                                       & x                         & T                                                             & x     \\
      & 2020          & Prediction uncertainty~\cite{Passive-Active-WB-MIA}                                            & Update-based $\dagger$ $\star$                 & DL                                                               & \Circle                                                 & A \& P                                             & x                         & T \& I                                                        & x     \\
                                                                     & 2021          & User-level MIA~\cite{zhao2021user}                                                    & Update-based $\dagger$ $\star$                   & DL                                                               & \Circle                                                 & P                                   & x                         & I                                                             & x \\
                                                                     & 2021          & Trend correction~\cite{efficient-trend-based-fl}                                                  & Trend-based $\star$                                           & DL                                                               & \Circle                                                 & P                    & x                         & T \& I                                                        & \checkmark     \\
                                                                     & 2021          & SIA~\cite{source-inf-fl}                                                               & Trend-based                                                                & DL                                                               & \Circle                                                 & P                                           & \checkmark & T \& I                                                        & \checkmark     \\
                                                                     & 2022          & CS-MIA ~\cite{CS-MIA}                                                           & Trend-based  $\dagger$ $\star$                   & DL                                                               & \Circle                                                 & A \& P                                                & \checkmark & T \& I                                                        & x \\
                                                                     & 2023          & Local Bias and Global Bias~\cite{MIA-FL-Bias}                                        & Trend-based  $\dagger$ $\star$                   & DL                                                               & \Circle                                                 & P                                                 & \checkmark & T \& I                                                        & x     \\
                                                                     & 2023          & GBMIA~\cite{wang2023gbmia}                                                             & Update-based $\dagger$ $\star$                   & DL                                                               & \Circle                                                 & A                                 & x                         & T \& I                                                        & x \\
                                                                     & 2023          & MIA-BAD~\cite{MIA-BAD-FL}                                                           & Local shadow model-based $\dagger$                                                  & DL                                                               & \CIRCLE                                                 & P                                                     & x                         & I                                                             & x \\ \bottomrule
\end{tabular}
}
{\footnotesize
$\dagger$ and $\star$  corresponds to local and global attack respectively. \CIRCLE  represents the black-box setting, \Circle the white-box and \LEFTcircle both.
}
\caption{Proposals of Membership Inference Attacks for centralized and federated learning paradigms.}
\label{tab:summary}
\end{table*}

Despite the extensive literature on MIAs, we identify the following important gaps: \textit{i)} most proposals, while reporting results on both data types, tend to focus on tabular datasets with hundreds of labels which require complex learning models and make evaluations on tabular datasets unreliable~\cite{Systematic-eval, Adv-Regul,Passive-Active-WB-MIA,White-Box-MIA-DL,shadow-update-based-fl,efficient-trend-based-fl,MIA-FL-Bias}; \textit{ii)}
the robustness of many attacks has not been tested against defense mechanisms, potentially overestimating their effectiveness in well-generalized models; \textit{iii)}
in FL settings, evaluations typically assume an insider adversary, such as a server or participant, whereas an outsider adversary scenario remains underexplored; \textit{iv)} prior work has shown that rare data points, such as outliers, increase vulnerability to MIAs~\cite{Long-MIA-Well-Gen}, however the effect of single-outs, which are unique patterns as well, has not yet been investigated, and \textit{v)} little attention has been given to the transferability of attacks, where adversaries use surrogate models with different architectures~\cite{Salem-ML-Leaks,MIA-in-MLaaS}.
We aim to address all these issues in the following sections. 

\section{Tabular MIAs under Centralized and Federated Learning}~\label{sec:results}
In this section, we review and discuss potential MIAs and defense solutions from the literature. We aim to answer the following three research questions.

\begin{enumerate}[label=\textbf{RQ\arabic*}:]
    \item How effective are MIAs in tabular data across different learning settings?
    \item To which extent are single-outs vulnerable to MIAs?
    \item Do MIAs trained on different surrogates transfer to target models? 
\end{enumerate}

Depending on the prior knowledge the adversary has about $f_{Target}$, they can proceed with different attack strategies. We define two adversaries in which the assumptions of the second adversary are relaxed. 

\textit{Adversary 1.} Has additional knowledge, such as the architecture of $f_{Target}$ and the distribution of $D_{Target}^{Train}$.

\textit{Adversary 2.} Has knowledge of the data distribution but no access to the $f_{Target}$ model.

The first adversary assumption is used to test MIAs in different learning settings, including single-out evaluation. The second assumption will provide insights into attack transferability. In both situations, we consider the access to the target output vector.

\subsection{MIAs efficacy on different settings}
We test two main MIAs approaches: shadow model-based and reference model-based attacks. In the former, we demonstrate the shadow via predictions (predicted label)~\cite{Shokri-MIA-Against-ML} and via probabilities (top $k$ probabilities, where $k=2$ if the task is binary classification, otherwise $k=3$)~\cite{Salem-ML-Leaks}. In the latter, we apply the online and offline versions of LiRA~\cite{MIA-first-prin} and RMIA~\cite{rmia}. While online trains reference models separately for each target data (test query $x$), in which $n$ IN models are trained containing $x$ in their training set, offline uses only $n$ trained reference models on randomly sampled datasets, avoiding any training on test queries. 
We compare three learning scenarios: \textit{i)} MIAs in centralized with a weak grid search, \textit{ii)} centralized with defenses, and \textit{iii)} FL as defense strategy. 
We report the results of attacks using AUC. 
Further details about the experimental setup are described in the Appendix~\ref{app:experiments}.

\begin{figure}[!htb]
     \centering
     \begin{subfigure}[b]{\columnwidth}
         \centering
    \includegraphics[width=\columnwidth]{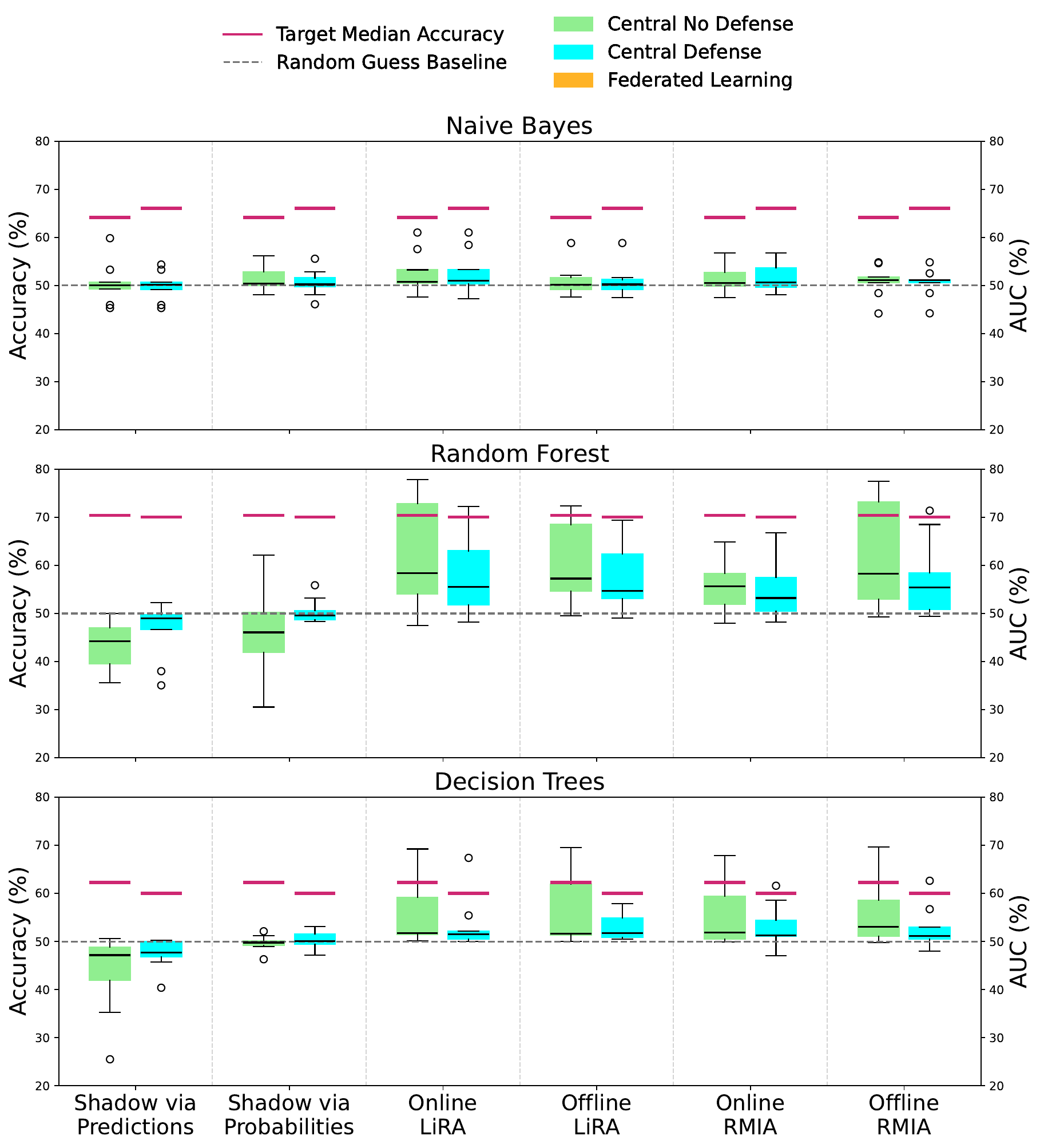}
         \caption{MIAs on centralized learning with and without defenses.}
         \label{fig:cl}
     \end{subfigure}
     \vfill
     \begin{subfigure}[b]{\columnwidth}
         \centering
         \includegraphics[width=\columnwidth]{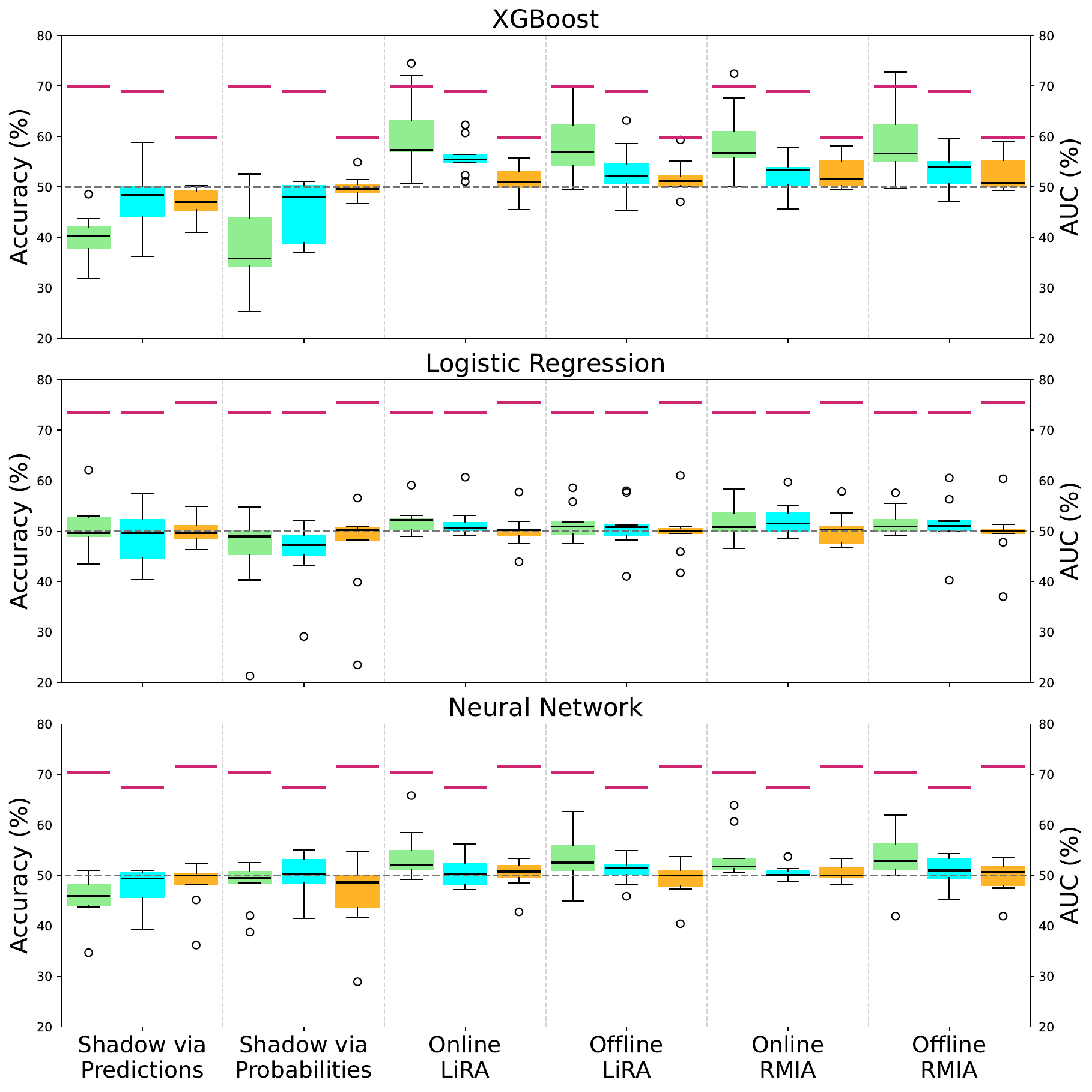}
         \caption{MIAs on centralized and federated learning.}
         \label{fig:fl}
     \end{subfigure}

        \caption{MIAs efficacy tested on ten datasets and six machine learning models for different learning paradigms.}
        \label{fig:overall_auc}
\end{figure}

The first set of results concerns the efficacy of MIAs across different learning paradigms. Considering the knowledge of \textit{Adversary 1}, Figure~\ref{fig:overall_auc} shows the variability of each MIA for nine datasets. Figure~\ref{fig:cl} compares their efficacy in centralized with and without defenses, while Figure~\ref{fig:fl} includes FL. Note that the first three models are not supported in the current available FL tools. 


Random Forest (RF), XGBoost and Decision Trees (DT) are clearly more vulnerable to LiRA and RMIA attacks with the first two presenting a median on centralized learning higher than 55\%. On the other hand, Naive Bayes (NB), Logistic Regression (LR), and Neural Network (NN) are the most robust models against MIAs regardless of the attack approach. 

For shadow model-based attacks, the use of prediction vectors or predicted labels, results in comparable performance, except for RF and DT. For the majority models, the marginal differences suggest that the additional information provided by prediction vectors offers limited advantage over label-only outputs in this setting, contributing only minimally to attack success.

Concerning reference model-based approaches, we notice a positive performance compared with shadow model-based. Furthermore, we observe that LiRA is marginally better than RMIA. However, the differences between them are not substantial. In addition, there is no clear difference between the online and offline versions. 
This makes it difficult to draw solid conclusions about the comparative behavior of LiRA and RMIA, particularly across their online and offline versions. Such poor performances may be due to the fact that online attacks require a considerable amount of target samples to tune on.

In centralized learning, incorporating defenses does not necessarily lead to stronger resilience against MIAs. 
We observe that shadow model-based tend to behave better when using hyper-parameter tuning as a defense. 
Nevertheless, we note that the median is very close to the random guess. On the other hand, LiRA and RMIA slightly reduce their attack efficacy, in which a higher impact is noted in RF and XGBoost. We also observe that hyper-parameter tuning does not affect the target accuracy by much. For NB in particular, the generalization is enhanced, leading to improved target accuracy, while other models achieve comparable accuracy. For NN, the target accuracy is considerably reduced as we combine early stopping and dropout. 

Regarding FL, Figure~\ref{fig:fl} shows that such strategy can be specially leveraged as a defense when using XGBoost. However, this model provides poor predictive performance results.
Conversely, this is not verified for the other two models. FL with LR or NN performs better than centralized while maintaining MIAs at low levels. 

\subsection{MIAs on single outs}
We provide in this section an analysis of the potential vulnerability of single-outs to MIAs. Single-outs are records that can be uniquely identified if an adversary knows the values of a specific set of quasi-identifiers. 
These are attributes that, when combined, generate a unique signature that may lead to re-identification.
For instance, date of birth, gender, geographical location, profession and ethnic group. $K$-anonymity is a method that allows us to identify single-outs~\cite{sweeney2002k}. 
Each equivalence class (group of records sharing the same characteristics) is assigned a frequency $f_k$. A record is unique when $f_k=1$. 

Given the extensive experiments and for the sake of conciseness, we present our results for centralized learning without defenses and one learning model (XGBoost). Table~\ref{tab:singleouts} shows the percentage of detected single-outs by each attack approach. We consider three datasets, in which the quasi-identifiers are well-defined (details in section~\ref{app:setup}).

\begin{table}[!ht]
\scriptsize
\resizebox{\columnwidth}{!}{%
\begin{tabular}{c|ccc}
\toprule
\textbf{Attack}          & \textbf{Covid} & \textbf{Dropout Success} & \textbf{Half Million} \\ \hline
Shadow via Predictions   & 99.7           & 62.5                     & 74.25                 \\
Shadow via Probabilities & 76.67          & 62.5                     & 57.77                 \\
Online LiRA              & 52.42          & 87.5                     & 68.6                  \\
Offline LiRA             & 62.42          & 62.5                     & 58.08                 \\
Online RMIA              & 43.33          & 100.0                    & 100.0                 \\
Offline RMIA             & 60.91          & 100.0                    & 99.69 \\
\bottomrule
\end{tabular}%
}

\caption{Percentage of single-outs captured by each attack in centralized learning and XGBoost.}
\label{tab:singleouts}
\end{table}

Results demonstrate a clear vulnerability of XGBoost regarding single-outs. While shadow model-based attacks perform well in \textit{Covid}, RMIA is stronger in \textit{Dropout Success} and \textit{Half Million}. The single outs of \textit{Dropout Success} are the most sensitive to all attacks. 
These results also enforce that the large variation across datasets reveals that MIAs on single-outs are data-dependent, influenced by the granularity of the quasi-identifiers.


\subsection{Transferability of MIAs}
So far, we have considered a white-box setting. However, an adversary might have less information, as described in  \textit{Adversary 2}. In this section, we investigate whether a MIA trained using different surrogate transfers remains effective against a different target model. Table~\ref{tab:transfer} presents the AUC of each combination for centralized learning (no defenses) and using the \textit{Cirrhosis} dataset. 

\begin{table}[!ht]
\scriptsize
\resizebox{\columnwidth}{!}{%
\renewcommand{\arraystretch}{1}
\begin{tabular}{c|c|cccccc}
\toprule 
\multirow{2}{*}{\textbf{Attacks}}                                                   & \multirow{2}{*}{\textbf{Target Model}} & \multicolumn{6}{c}{\textbf{Attack Model}}                                                                                \\ \cline{3-8} 
                                                                                    &                                        & \textbf{NN}                   & \textbf{NB}                   & \textbf{DT}                   & \textbf{RF}                   & \textbf{XGB}                  & \textbf{LR}             \\ \hline
                                                                                    \multirow{6}{*}{\begin{tabular}[c]{@{}c@{}}Shadow via \\ Predictions\end{tabular}}  & NN                                     & {\ul 48.31}          & 54.43                & 45.75                & 47.1                 & 46.65                & \textbf{53.58} \\
                                                                                    & NB                                     & 47.69                & {\ul \textbf{59.83}} & 44.09                & 43.56                & 44.64                & 55.7           \\
                                                                                    & DT                                     & 39.14                & \textbf{64.74}       & {\ul 35.21}          & 37.26                & 36.24                & 54.41          \\
                                                                                    & RF                                     & 47.56                & \textbf{58.6}        & 43.29                & {\ul 44.78}          & 44.86                & 56.21          \\
                                                                                    & XGB                                    & 43.72                & \textbf{65.18}       & 37.17                & 39.23                & {\ul 39.44}          & 58.17          \\
                                                                                    & LR                                     & 45.18                & \textbf{52.15}       & 44.73                & 44.86                & 46.49                & {\ul 40.35}    \\ \hline
\multirow{6}{*}{\begin{tabular}[c]{@{}c@{}}Shadow via\\ Probabilities\end{tabular}} & NN                                     & {\ul 50.77}          & 47.3                 & 43.93                & \textbf{54.73}       & 53.47                & 40.62          \\
                                                                                    & NB                                     & 55.87                & {\ul 56.16}          & 50.66                & 44.71                & 50.84                & \textbf{56.98} \\
                                                                                    & DT                                     & 46.12                & 47.61                & {\ul 52.12}          & 48.08                & \textbf{52.24}       & 48.01          \\
                                                                                    & RF                                     & \textbf{58.45}       & 42.61                & 36.44                & {\ul 56.71}          & 43.11                & 37.77          \\
                                                                                    & XGB                                    & \textbf{57.96}       & 48.77                & 41.51                & 52.62                & {\ul 43.79}          & 45.03          \\
                                                                                    & LR                                     & 50.82                & 40.53                & 44.45                & \textbf{57.1}        & 47.26                & {\ul 40.35}    \\ \hline
\multirow{6}{*}{\begin{tabular}[c]{@{}c@{}}Online\\ LiRA\end{tabular}}              & NN                                     & {\ul 58.54}          & 48.48                & 55.33                & \textbf{58.73}       & 52.18                & 39.2           \\
                                                                                    & NB                                     & 47.65                & {\ul \textbf{57.56}} & 56.99                & 47.97                & 49.46                & 49.43          \\
                                                                                    & DT                                     & 36.13                & 49.09                & {\ul \textbf{59.21}} & 41.61                & 44.18                & 45.75          \\
                                                                                    & RF                                     & 58.51                & 50.15                & 56.55                & {\ul \textbf{62.72}} & 56.18                & 51.18          \\
                                                                                    & XGB                                    & 55.14                & 58.91                & 57.62                & \textbf{63.5}        & {\ul 62.54}          & 54.22          \\
                                                                                    & LR                                     & \textbf{58.28}       & 53.68                & 57.06                & 56.47                & 55.14                & {\ul 51.81}    \\ \hline
\multirow{6}{*}{\begin{tabular}[c]{@{}c@{}}Offline\\ LiRA\end{tabular}}             & NN                                     & {\ul \textbf{55.88}} & 44.86                & 50.3                 & 54.96                & 50.07                & 47.3           \\
                                                                                    & NB                                     & 55.49                & {\ul 52.11}          & 54.84                & \textbf{56.25}       & 56.05                & 53.94          \\
                                                                                    & DT                                     & 56.8                 & 60.61                & {\ul 61.95}          & 62.2                 & 61.11                & \textbf{63.54} \\
                                                                                    & RF                                     & 49.87                & 44.16                & 49.67                & {\ul \textbf{60.65}} & 52.26                & 49.33          \\
                                                                                    & XGB                                    & 59.47                & 48.48                & 53.25                & \textbf{63.31}       & {\ul 62.35}          & 56.92          \\
                                                                                    & LR                                     & 58.65                & 49.52                & 52.63                & \textbf{60.69}       & 57.4                 & {\ul 58.62}    \\ \hline
\multirow{6}{*}{\begin{tabular}[c]{@{}c@{}}Online\\ RMIA\end{tabular}}              & NN                                     & {\ul 51.63}          & \textbf{55.51}       & 48.54                & 48.06                & 52.74                & 49.63          \\
                                                                                    & NB                                     & 58.3                 & {\ul 52.74}          & 54.86                & 57.62                & 55.53                & \textbf{59.01} \\
                                                                                    & DT                                     & 58.47                & 54.53                & {\ul 59.39}          & \textbf{60.61}       & 60.11                & 59.97          \\
                                                                                    & RF                                     & 52.55                & 46.73                & 56.86                & {\ul \textbf{58.34}} & 55.92                & 51.68          \\
                                                                                    & XGB                                    & 58.84                & 55.05                & 58.76                & 60.26                & {\ul \textbf{61.02}} & 60.48          \\
                                                                                    & LR                                     & 58.36                & 51.33                & 58.08                & \textbf{60.93}       & 59.63                & {\ul 53.96}    \\ \hline
\multirow{6}{*}{\begin{tabular}[c]{@{}c@{}}Offline\\ RMIA\end{tabular}}             & NN                                     & {\ul 56.27}          & 48.22                & 54.36                & \textbf{57.69}       & 51.2                 & 49.19          \\
                                                                                    & NB                                     & \textbf{58.04}       & {\ul 54.64}          & 54.73                & 57.93                & 54.94                & 56.18          \\
                                                                                    & DT                                     & 60.39                & 58.76                & {\ul 58.54}          & \textbf{62.46}       & 59.34                & 59.5           \\
                                                                                    & RF                                     & 57.21                & 50.98                & 59.84                & {\ul \textbf{60.47}} & 57.32                & 56.58          \\
                                                                                    & XGB                                    & 62.94                & 58.45                & 62.98                & \textbf{65.88}       & {\ul 62.44}          & 62.67          \\
                                                                                    & LR                                     & 59.36                & 53.94                & 61.76                & \textbf{62.98}       & 56.43                & {\ul 57.62}    \\ \bottomrule

\end{tabular}%
}
\caption{Performance (AUC) of transferring MIAs in centralized learning using \textit{Cirrhosis} dataset. The baseline is underlined, and the best attack model is in bold.}
\label{tab:transfer}
\end{table}  

For this particular dataset, we observe that a large portion of the combinations perform better when using different surrogate models. In particular, we observe that RF is more accurate for several settings. This may be related to the characteristics of the inference set and the model. RF is known for its stability, so we obtain more stable probabilities making it easier for an attack model to learn the membership confidence levels.
However, for shadow via predictions, NB is the best candidate, achieving a considerably higher AUC compared to the previous lower values obtained with the same model. 

\section{Discussion}~\label{sec:discussion} 
\textbf{On the poor performance of MIAs model-based.}
To better understand the different factors that can influence the effectiveness of MIAs in tabular data, in Appendix~\ref{app:more_results} we provide Table~\ref{tab:tprfpr} with averaged AUC and TPR@1\%FPR of MIAs across learning models for each dataset for centralized learning.
When considering AUC, we verify that Online LiRA is dominant across datasets. Yet, other reference model-based are more efficient in capturing members when FPR is fixed at 1\%. Also, for some datasets, there is high variability, which means that certain models may be more vulnerable. Nevertheless, this corroborates previous results (Figure~\ref{fig:overall_auc}), in which MIAs based on reference models are substantially more effective than shadow model attacks, making them the dominant threat model.  
However, the overall results demonstrate poor MIAs performance (\textbf{RQ1}): marginally better than random guess.
This contrasts with prior observations in the literature even for overfitted models. 

Moreover, we identify interesting isolated cases: for example, the RF trained on the \textit{Locations} dataset exhibits a generalization gap of 41\% but achieves an AUC of 50.69\% under the Online LiRA attack. On the other hand, the \textit{Dropout Success} dataset shows a smaller generalization gap (12.6\%) yet a relatively high AUC of nearly 60\%.
These observations suggest that the relationship between overfitting and vulnerability to MIAs is not strictly linear but also emphasize that, although well-generalized models generally offer stronger privacy guarantees, they can still reveal membership information under specific conditions. Similar conclusions have been previously reported in the image data domain~\cite{Yeom, sablayrolles2019white}.

\textbf{Centralized vs FL.}
Our analysis shows that, for the best-performing attacks, applying hyper-parameter tuning and regularization techniques mitigates the attack's efficacy (Figure~\ref{fig:cl}). In particular, hyper-parameter tuning which helps find models that generalize better, demonstrates that it is possible to achieve a favorable trade-off between privacy and predictive performance. 

Concerning FL, our results show that even with such limited knowledge (outsider attack), MIAs are possible for certain configurations (Figure~\ref{fig:fl}). However, most of the attack strategies demonstrate a performance close to the random guess. 
Hence, FL not only allows secure data sharing, but can also improve model accuracy with lower MIAs success for specific settings. 
Note that when building an FL system, it is important to focus on weights updates rather than gradients. For instance, FedAvg uses weights updates, while FedSGD uses raw gradients~\cite{CS-MIA}. The latter may be more vulnerable to MIAs as it depends on the raw gradients sharing. 

\textbf{The influence of single-outs.}
Despite of the generally low performance of MIAs in tabular data, we found that single-outs contribute to the attack's performance (\textbf{RQ2}). 

These rare data points must be protected, either through efficient model defense (in-processing) or directly within the data (pre-processing). Concerning the former, it is important to re-evaluate MIA performance after applying a defense strategy, as certain attacks may exhibit resistance to particular defenses, thereby exposing more single-outs. 
Pre-processing involves transforming the data by reducing its granularity or adding noise to the quasi-identifiers. Several studies have demonstrated the effectiveness of such methods in maintaining data utility while preserving privacy~\cite{carvalho2025differentially,carvalho2025empowering}. Finally, it is crucial to consider different attack scenarios by varying the set of quasi-identifiers, since it is difficult to know which information an adversary might possess. 


\textbf{Impact of adversary knowledge.}
Due to the phenomenon of attack transferability, adversaries can exploit the fact that attacks crafted on one model often transfer to another trained on similar data even without direct access to the target model~\cite{MIA-in-MLaaS}. Most of the relevant literature lacks analysis of model attack transferability as they typically assume knowledge of the target model. 

We observed enhanced MIAs when the surrogates are different from the target (\textbf{RQ3}). This highlights the feasibility of conducting improved MIAs without prior knowledge of the target model. However, this may require several trials until the best attack model is achieved.

\begin{figure}[!htb]
      \centering
      \begin{subfigure}[b]{\columnwidth}
         \centerline{\includegraphics[width=\columnwidth]{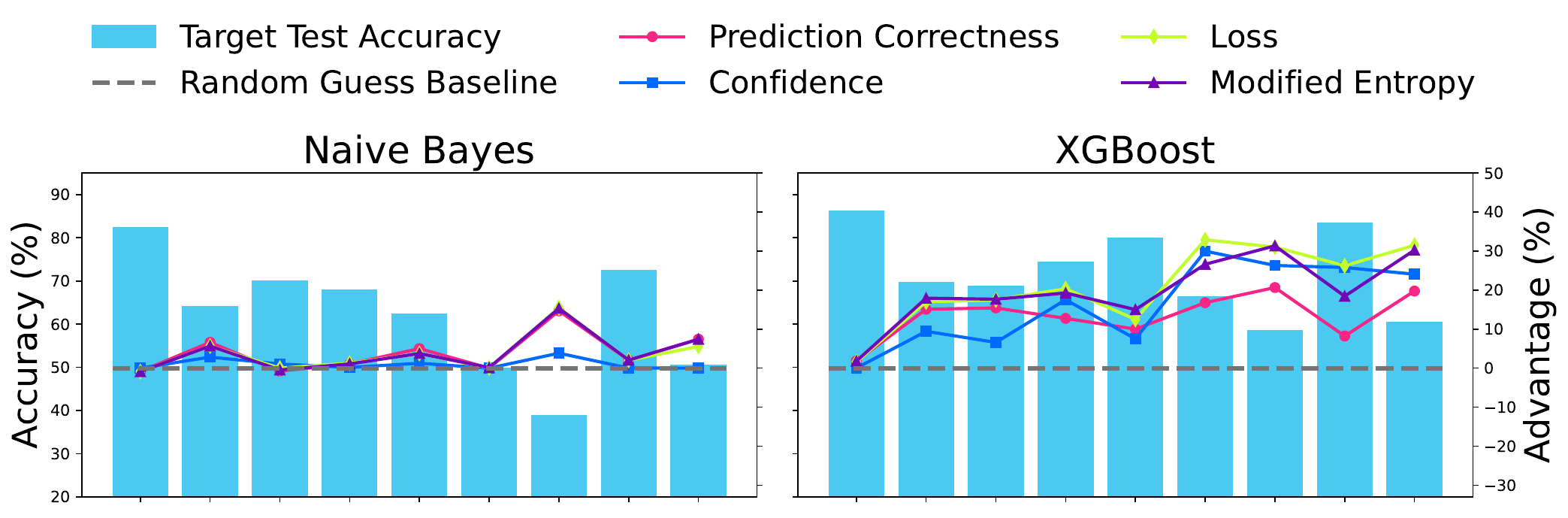}}
      \end{subfigure}
      \vfill
      \begin{subfigure}[b]{\columnwidth}
          \centerline{\includegraphics[width=\columnwidth]{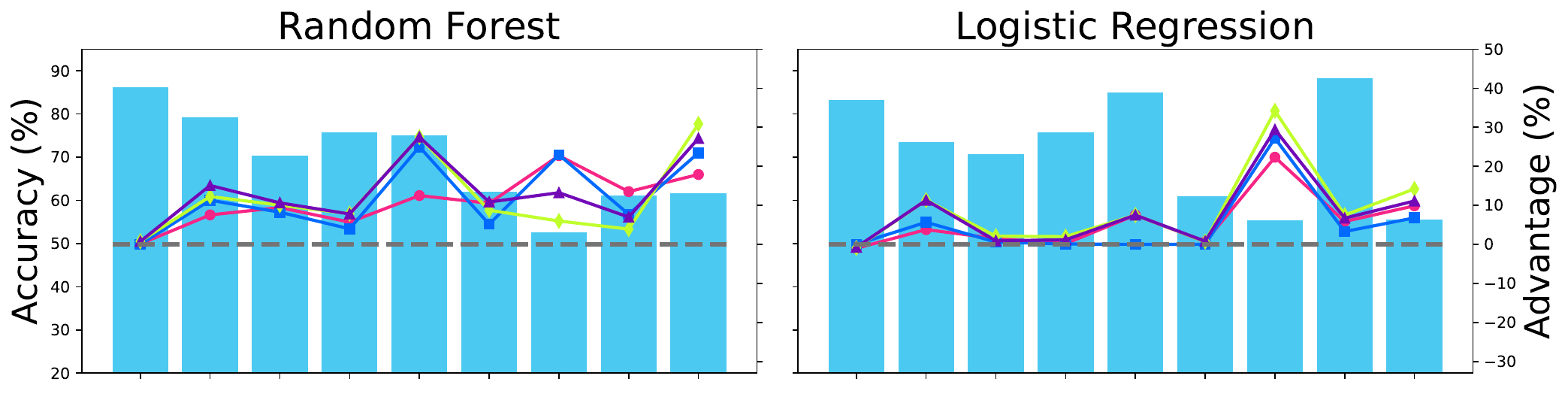}}
      \end{subfigure}
      \vfill
      \begin{subfigure}[b]{\columnwidth}
          \centerline{\includegraphics[width=\columnwidth]{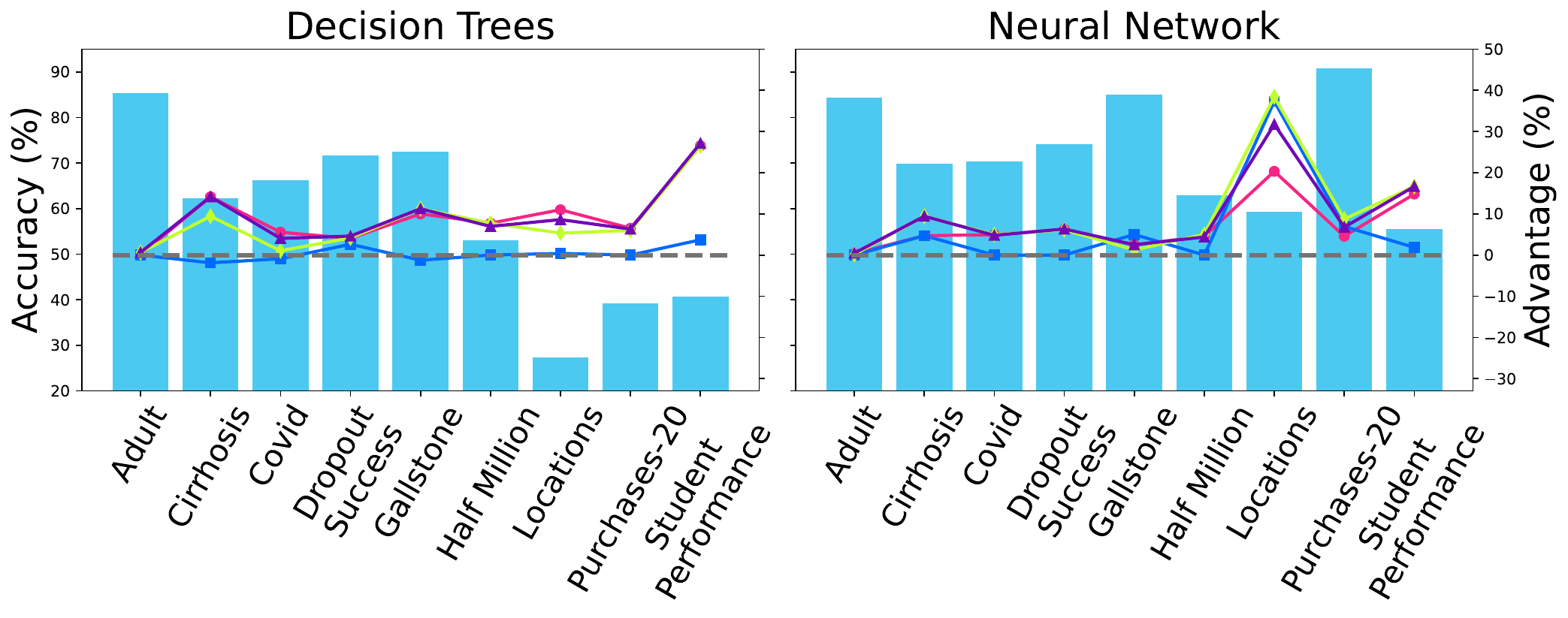}}
      \end{subfigure}
         \caption{MIA Advantage using metric-based attacks on the centralized learning paradigm.}
         \label{fig:roc_auc}
         \label{fig:metric-based}
 \end{figure}

\textbf{The fragility of metric-based approaches.}
We also tested whether an attack model-free approach, namely metric-based, can be a solution. We depict in Figure~\ref{fig:metric-based} the centralized paradigm without defenses. In this case, it is not possible to obtain AUC at several thresholds given the nature of the prediction correctness attack. Thus, we report the MIA Advantage (TPR-FPR), which measures how much better the attack is compared to random guessing.

In contrast with shadow- and reference model-based, these metric attacks present a slightly higher attack inference, especially for XGBoost with a median for each attack between 65\% and 70\%. 
There is a similar behavior between NB, LR and NN such as observed before. 

The observed peaks in the Advantage highlight the overfitting of the models. Conversely, when the model is well-generalized, the Advantage is near zero. This allows us to conclude that the effectiveness of metric-based approaches strongly depends on overfitting levels (Table~\ref{tab:target_acc} in Appendix~\ref{app:more_results} shows details for training and test accuracies for all target models).


\textbf{Attack value vs. attack cost.} In the context of MIAs, value can be characterized by attack accuracy, while the cost of attack
refers to the knowledge and resources needed for a successful attack. 

From a cost perspective, we employed attack model-based and attack model-free approaches. In the former, we used only one shadow model, as it has been proven to be sufficient to capture target characteristics~\cite{Salem-ML-Leaks,Adv-Regul}, and in fact, we verified that it produces predictive performance similar to the target model and the attack exhibits high accuracy (Appendix~\ref{app:more_results}, Figure~\ref{fig:xgb_acc}). This does not apply for reference model-based as it typically requires several models to analyze the impact of IN/OUT target point. Therefore, a shadow model-based on such a design is less costly than reference attacks. Yet, for enhanced attack, experiments should be conducted on different surrogate models, which obviously leads to more costs but at the same time does not require the knowledge of the target model. 
In contrast, metric-based, which does not require any attack model training, has demonstrated efficacy to a certain extent. However, a major drawback, is that their efficacy strongly depends on overfitting and cannot be used on ML as a Service (MLaaS) as it requires the true labels of the target model. Yet, it can be used before deploying ML models for previous auditing. 
A few methods have been proposed for auditing. For instance, the access to model's hyper-parameters~\cite{suri2024parameters, Hierarchical-tree-based} can help auditing performance attacks which can guide future developments. 

To conclude, our findings highlight that there is no one-size-fits-all solution for evaluating MIAs on tabular data. Although some learning configurations yield improved attack performance compared with random guess, no single attack approach consistently performs well across datasets and models. Furthermore, the level of overfitting is not always indicative of better MIAs.
This emphasizes the need for continuous auditing and the development of stronger MIAs evaluations.
We stress that weak attacks might underestimate true privacy inference, potentially creating a false sense of privacy.

\section{Open issues and research directions}
MIAs on tabular data require dealing with diverse column types and ambiguous semantics, which may result in inconsistent inference signals and more complex attack than with other data modalities. Such characteristics may introduce different patterns and consequently reduce the attack effectiveness. 
This does not happen in domains with more uniform structure, such as images. Thus, explainable AI methods, which reveal underlying patterns, may help to distinguish between members and non-members~\cite{lucieri2023translating}. 


Usually, the adversary is defined based on different levels of knowledge about the target model. However, a concern remains about the knowledge of the underlying data distribution. In MLaaS, an adversary may have even more limited access to data domain, leading to a distribution shift between the target and the surrogate model~\cite{rmia}. This mismatch not only undermines the reliability of MIAs but also implies that attack models may need to be continuously adjusted as the target model evolves or is updated with new data. Consequently, future research should explore adaptive attack strategies capable of maintaining effectiveness under dynamic target conditions that account distribution shift.

Inferring membership in VFL is not an easy task. Whereas an insider adversary in HFL controls a entire local model and has access to gradients of all its parameters, in VFL, the adversarial participant only controls part of the federated model, which cannot operate independently, and thus only has access to the gradients of this incomplete model. 
In VFL, participants need to obtain the coincident sample space, thus the MIA from other participants may be redundant~\cite{wei2022vertical}. 
Since there is only one participant owning the labels, an adversary can infer the private labels~\cite{arazzi2025defense,fu2022label}. 
Nevertheless, we emphasise that MIAs can be executed on the global model through an outsider attack, regardless of the setting of FL. 

Recent studies have explored enhancing MIAs through poisoning attacks either in centralized~\cite{chen2022amplifying} and FL~\cite{he2024enhance}, as well as through adversarial attacks~\cite{MemGuard,li2024privacy}. However, this may fall short when targeting well-generalized models. In MLaaS, models are expected to be well-generalized due to controlled deployment and monitoring that ensure reliable performance. Moreover, practitioners often adopt anti-overfitting techniques to maintain users' trust. Consequently, it becomes crucial to develop stronger privacy evaluation that do not rely solely on overfitting as an indicator of vulnerability.
 
The foundations of MIAs can be transferred to training data reconstruction attacks. Rather than determining whether a data point was included in the training set, these attacks aim to recover specific information about it. The adversary tries to reconstruct one record among $n$ possible candidates, where $n$ reflects their prior knowledge they have. Defenses such as differential privacy against MIAs can be leveraged for reconstruction attacks, in which the parameter $\epsilon$ can provide enough information to characterize vulnerability against reconstruction attacks~\cite{hayes2023bounding}.

In light of the rapid technological advancements that are continuously enhancing the capabilities of adversaries, it is essential to implement privacy mechanisms prior to model training. A dual-layer privacy-preserving approach should be considered in AI-based applications, whether in centralized or FL paradigms. It has been proven that safeguarding input data privacy does not necessarily compromise predictive performance in the centralized setting~\cite{carvalho2025differentially}. Similar efforts are emerging in FL, where anonymization techniques such as generalization or microaggregation have been proposed~\cite{huang2023dual}. Hence, strengthening input data protection (pre-processing) may prevent further attacks on learning models with no impact on target predictive performance.

\section{Conclusion}~\label{sec:conclusion}
In this paper, we present a systematization of knowledge for tabular MIAs in centralized and federated learning. We update and suggest a new classification taxonomy for both settings. We also explore defense strategies. Then, we evaluate different MIAs strategies (attack model-based and attack model-free) across several tabular datasets and learning models that focus on solving simple tasks.
Contrary to the often reported success of MIAs, our experiments reveal limited attack performance across various models and defense strategies with tabular data.
Moreover, we investigate the vulnerability of single-outs to MIAs as learning models often tend to memorize rare data points. Our results demonstrate that even MIAs with low attack performance can capture high percentages of such sensitive data points.
Lastly, we study the transferability of MIAs, and show that attacks are more effective when trained on surrogate models that differ from the target. 
In short, we underscore the continuous efforts in developing more robust MIAs that allow for auditing in tabular data. 

For future work, we plan to investigate how MIAs can be enhanced with data augmentation in tabular data.
Also, in terms of defenses, it is important to understand the efficacy of pruning NN~\cite{ijcai2021p432} and the influence of the number of rounds and nodes in FL.


\bibliography{Bibliography}
\bibliographystyle{ieeetr}




\appendices
\section{Experimental Details}\label{app:experiments}
This section describes the full experimental setup of the implemented MIAs and respective defenses. 

\subsection{Data}\label{app:data}

The characteristics of the nine datasets used to evaluate the MIAs are presented in Table~\ref{tab:data}. We use datasets that span a range of sample sizes, features, and labels. The domains include healthcare, finance, education, locations and purchases. 

\begin{table}[ht!]
\scriptsize
\resizebox{\columnwidth}{!}{%
\begin{tabular}{l|lcccl}
\toprule
\textbf{Dataset}                                              & \textbf{Domain} & \textbf{\# Rows} & \textbf{\# Features} & \textbf{\# Labels} & \multicolumn{1}{c}{\textbf{Classification task}}         \\ \hline
Adult~\cite{ds:adult}                                                        & Finance         & 29096            & 13                   & 2                  & Annual income of individuals exceeds \$50K/yr \\
Cirrhosis~\cite{ds:cirrhosis}                                                     & Healthcare      & 417              & 17                   & 3                  & Survival state of patients with liver cirrhosis  \\
Covid~\cite{carvalho2025empowering}                                                         & Healthcare      & 2671             & 29                   & 3                  & Outcome of patient with covid-19                 \\
Dropout Success~\cite{ds:dropout_success}  & Education       & 4423             & 36                   & 3                  & Student dropout and academic success             \\
Half  Million~\cite{ds:half_million}        & Finance         & 15855            & 28                   & 12                 & Lifestyle category of individuals                \\
Gallstone~\cite{ds:gallstone}                                                     & Healthcare      & 318              & 38                   & 2                  & Presence or absence of gallstones in a patient   \\
Locations\cite{ds:Shokri_Datasets,ds:location-unprocessed}                                                     & Locations       & 5010             & 446                  & 30                 & Geosocial type of a user                         \\
Purchases-20~\cite{ds:Shokri_Datasets,ds:purchases-unprocessed}                                                  & Purchases       & 48446            & 600                  & 20                 & Purchasing style of a user                       \\
\begin{tabular}[c]{@{}l@{}}Student\\ Performance~\cite{ds:student_performance}\end{tabular} & Education       & 648              & 30                   & 3                  & Student achievements in secondary education      \\
\bottomrule
\end{tabular}%
}
\caption{Specifications of the evaluation datasets.}
\label{tab:data}
\end{table}

\subsection{Setup}\label{app:setup}


We use 50\% of the data to represent the population data.
The private data is splitted into 75\% train and 25\% test sets, and used to train and validate the target model. If the target model is trained via FL, then both sets are equally split between three distinct participants. In order to test the performance of the attacks, we create a membership inference set that is composed of an equal number of members and non-members.
Therefore, we only include 25\% of the records from the train set of the target model accompanied by all the records from the test set. The split of ground-truth members from this set is additionally used to evaluate the single-outs.

We evaluate the attacks on a simple Neural Network and on traditional ML models. The architectures of the target models vary depending on the specified training set configuration. For a target model trained locally without defenses, its Neural Network architecture includes two linear, fully connected layers with 128 neurons and one \textit{ReLU} activation function. In addition, an \textit{Adam} optimizer is used with a learning rate of 0.001 and a weight decay of \(1e-7\). A scheduler is added with a step size of 50 and gamma 0.1, and \textit{Cross-Entropy} loss is used for the loss function. The number of epochs and the batch size is 200. As a centralized defense, in order to make the model more robust against attacks, a dropout layer with probability 0.5 is added to the architecture of the previously described target model. Also, the \textit{Adam} optimizer learning rate is reduced from 0.001 to 0.0001. 

The remaining models require hyper-parameter tuning for each scenario. For the \textit{Central No Defense} scenario, a coarse and less precise grid search is defined. In contrast, in the \textit{Central Defense} scenario, a finer and more precise grid search is implemented. Table~\ref{tab:central_params} presents the hyper-parameter values used in the grid search of these scenarios.

\begin{table}[!ht]
\scriptsize
\resizebox{\columnwidth}{!}{%
\renewcommand{\arraystretch}{1}
\begin{tabular}{c|l|l|l}
\toprule
\textbf{Model}                                                                 & \multicolumn{1}{c|}{\textbf{Hyper-parameter}} & \multicolumn{1}{c|}{\textbf{Weak Grid Search}} & \multicolumn{1}{c}{\textbf{Defense Grid Search}}                                                \\ \hline
\multirow{2}{*}{\begin{tabular}[c]{@{}c@{}}Naive\\ Bayes\end{tabular}}         & \multirow{2}{*}{var\_smoothing}               & \(1e-9\),                                     & \(1e-9\),                                                                                       \\
                                                                               &                                               & np.logspace(0, -9, num=10)                    & \multicolumn{1}{c}{np.logspace(0, -9, num=100)}                                                 \\ \hline
\multirow{4}{*}{\begin{tabular}[c]{@{}c@{}}Decision\\ Trees\end{tabular}}      & criterion                                     & gini                                          & gini, entropy                                                                                   \\
                                                                               & max\_depth                                    & 6, 8, 10                                      & 2, 4, 6, 8                                                                                      \\
                                                                               & min\_samples\_leaf                            & 1, 2                                          & 3, 5, 7, 9                                                                                      \\
                                                                               & min\_samples\_split                           & 2                                             & 2, 5, 10                                                                                        \\ \hline
\multirow{5}{*}{\begin{tabular}[c]{@{}c@{}}Random\\ Forest\end{tabular}}       & max\_depth                                    & 6, 8, 10                                      & 2, 4, 6, 8                                                                                      \\
                                                                               & n\_estimators                                 & 100, 400                                      & 300, 500, 700                                                                                   \\
                                                                               & min\_samples\_leaf                            & 1, 2                                          & 3, 5, 7, 9                                                                                      \\
                                                                               & min\_samples\_split                           & 2                                             & 2, 5, 10                                                                                        \\
                                                                               & max\_features                                 & None                                          & 0.3, 0.6, 0.9                                                                                   \\ \hline
\multirow{9}{*}{XGBoost}                                                       & max\_depth                                    & 6, 8, 10                                      & 2, 4, 6, 8                                                                                      \\
                                                                               & \multirow{2}{*}{n\_estimators}                & \multirow{2}{*}{100, 400}                     & \multirow{2}{*}{\begin{tabular}[c]{@{}l@{}}1000 with 5 rounds\\    of early stopping\end{tabular}} \\
                                                                               &                                               &                                               &                                                                                                 \\
                                                                               & min\_child\_weight                            & 4, 7                                          & 5, 7, 9                                                                                         \\
                                                                               & subsample                                     & 1                                             & 0.3, 0.6                                                                                        \\
                                                                               & colsample\_bytree                             & 1                                             & 0.3, 0.6                                                                                        \\
                                                                               & learning\_rate                                & 0.3                                           & 0.001, 0.01, 0.1                                                                                \\
                                                                               & gamma                                         & 0                                             & 0, 1                                                                                            \\
                                                                               & reg\_alpha                                    & 0                                             & 0, 1                                                                                            \\
                                                                               & reg\_lambda                                   & 1                                             & 10                                                                                              \\ \hline
\multirow{5}{*}{\begin{tabular}[c]{@{}c@{}}Logistic\\ Regression\end{tabular}} & penalty                                       & None                                          & l2, elasticnet                                                                                  \\
                                                                               & l1\_ratio                                     & None                                          & None, 0.3, 0.5, 0.7                                                                             \\
                                                                               & C                                             & 0.0                                           & 0.0001, 0.001, 0.01, 0.1, 1                                                                     \\
                                                                               & solver                                        & lbfgs                                         & lbfgs, newton-cg, saga, sag                                                                     \\
                                                                               & tol                                           & \(1e-5\), \(1e-4\)                            & \(1e-5\), \(1e-4\), 0.001, 0.01                                                                 \\ \bottomrule
\end{tabular}%
}
\caption{Hyper-parameter grid search values for the centralized learning paradigm with and without defenses.}
\label{tab:central_params}
\end{table}

The FL system is built using NVFLARE~\cite{nvflare}. In this setting, Neural Network, XGBoost and Logistic Regression models follow the same architecture as the one from the centralized without defenses. The remaining models are currently not supported in NVFLARE, thus not implemented in this environment.





Finally, concerning single outs we only consider datasets for which the quasi-identifiers are generally known. For \textit{Covid} dataset we select \textit{Age, DateOfFirstPositiveLabResult, HospitalisationDays} and \textit{Gender} which results in 330 single outs. In dataset \textit{Dropout Success} we use \textit{Daytime/evening attendance, Nationality, Educational special needs, Gender} and \textit{International} QIs which leads to 8 single outs. Lastly, selecting  \textit{Gender, State, Country} and \textit{Age} in \textit{Half Million} we obtained 1856 single outs.


\subsection{Attacks}\label{app:methods}

We evaluate MIA on six distinct attack methods, namely Shadow via Predictions, Shadow via Probabilities, Online and Offline LiRA, and Online and Offline RMIA.
For the shadow-based methods we train one single model since it has been proven efficient~\cite{Salem-ML-Leaks}. 
In contrast, for the remaining reference model-based methods, multiple models are trained. Specifically, most datasets follow the number of reference models specified in existing literature, that is 256 for an online attack and 128 for an offline attack. However, the \textit{Adult}, \textit{Half Million} and \textit{Purchases-20} datasets are trained with 64 models in the online setting and 32 in offline because of their large dimension and computational resource limitations. In the online attack, approximately half of the models include the target record while the other half does not. In contrast, the offline attack simply trains reference models on randomly sampled records from the population data.


In addition to shadow- and reference model-based, we also included metric-based strategies to performe MIAs. These methods rely solely on the probability vector output of the target model and do not require any additional assumptions or auxiliary models. 
We use prediction correctness, prediction loss, prediction confidence and modified entropy~\cite{MIA-ML-Survey}. All the calculations for each are as follows.

\textbf{Prediction correctness~\cite{Yeom}.}
\begin{equation}~\label{eq:pred_corr}
    \mathcal{M}_{\mathrm{corr}}(\hat{p}(y \mid x), y)=\mathbbm{1}(\operatorname{argmax} \hat{p}(y \mid x)=y)
\end{equation}

\textbf{Prediction loss~\cite{Yeom}.}
\begin{equation}
    \mathcal{L}(\hat{p}(y \mid x) ; y) = -\sum_{i} y_{i} \log \left(p_{i}\right),
\end{equation}

where \(p_i\) is the confidence score in \((\hat{p}(y \mid x)\). Then, to infer membership, the following formula is used:

\begin{equation}~\label{eq:loss}
    \mathcal{M}_{\text {loss }}(\hat{p}(y \mid x), y)=\mathbbm{1}(\mathcal{L}(\hat{p}(y \mid x) ; y) \leq \beta),
\end{equation}

where $\mathcal{L}(\cdot)$ is the cross-entropy loss function and \(\beta\) corresponds to the threshold value. If its result is lower than the threshold, then the adversary classifies it as a member; otherwise, as a non-member.

\vspace{0.5em}
\textbf{Prediction confidence~\cite{Salem-ML-Leaks}.}

\begin{equation}~\label{eq:conf}
    \mathcal{M}_{\mathrm{conf}}(\hat{p}(y \mid x))=\mathbbm{1}(\max \hat{p}(y \mid x) \geq \beta).
\end{equation}





\textbf{Modified prediction entropy~\cite{Systematic-eval}.}

\begin{align}~\label{eq:mod_entr}
MH(\hat{p}(y \mid x), y)
  &= -\left(1 - p_y\right) \log\left(p_y\right) \notag\\
  &\quad - \sum_{i \neq y} p_i \log\left(1 - p_i\right)
\end{align}

where \(p_y\) is the confidence score of the ground-truth label. The attack is then defined as follows:

\begin{equation}~\label{eq:mod_entr_attack}
    \mathcal{M}_{\text {Mentr }}(\hat{p}(y \mid x), y)=\mathbbm{1}(M H(\hat{p}(y \mid x) ; y) \leq \beta).
\end{equation}

\vspace{0.5em}
For each of these metrics, the adversary has to be careful in choosing the threshold value, since this is a determinant factor in the membership inference result. Although an approach could be selecting random values, a three-step procedure is more efficient~\cite{Salem-ML-Leaks}. The first step would be to choose random points in the feature space of the target data point. Then, these random points would be queried in the target model to get the corresponding maximal posteriors and the respective top-$t$ percentile serves as the threshold. A shadow training technique can also be applied, in which different threshold values are obtained for different class labels of target models, such that the selected threshold is the one that achieves the highest accuracy in distinguishing between the shadow train and the test sets~\cite{Systematic-eval}.

\section{Complementary Results}\label{app:more_results}
In this section, we provide additional results and findings on the performance of MIAs on different attack methods and datasets.
\subsection{Decision tree defenses}
\begin{figure}[!ht]
\centerline{
\includegraphics[width=\columnwidth]{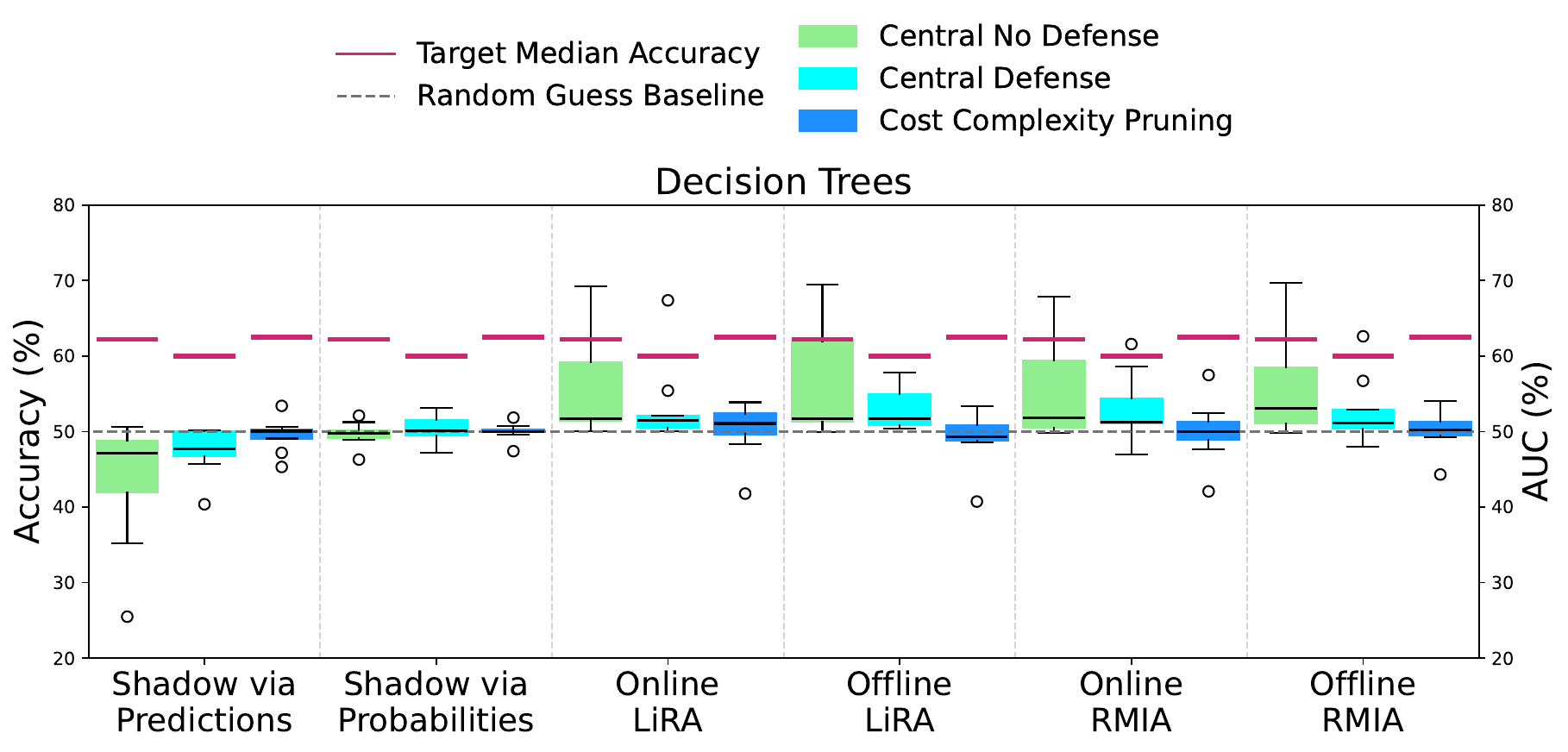}
}
\caption{Comparison of MIAs efficacy on Decision Trees for centralized learning without and with defense, including hyperparameter-tuning and pruning.}
\label{fig:dt_auc}
\end{figure}

Besides the hyper-parameter tuning, we also tested pruning the branches of DT that do not contribute to better performance as a defense strategy, namely cost-complexity pruning. Although pruning has not been tested in this context, it is a well-known strategy to reduce overfitting. Results are depicted in Figure~\ref{fig:dt_auc}. We observe a high efficacy in reducing the performance of the attack. However, this comes at the cost of target accuracy. 

\subsection{TPR@1\%FPR on centralized learning}
For an in-depth evaluation of the results for each dataset, we provide the performance of each attack method for all datasets considering TPR at 1\% FPR and AUC. The respective results presented in Table~\ref{tab:tprfpr} are averaged across all models and including the standard deviation.

\begin{table*}[!htb]
\scriptsize
\centering
\resizebox{\textwidth}{!}{%
\renewcommand{\arraystretch}{1.1}
\begin{tabular}{l|cc|cc|cc|cc|cc|cc}
\hline
\multicolumn{1}{c|}{\multirow{2}{*}{\textbf{Dataset}}} & \multicolumn{2}{c|}{\textbf{Shadow via Predictions}}  & \multicolumn{2}{c|}{\textbf{Shadow via Probabilities}} & \multicolumn{2}{c|}{\textbf{Online LiRA}}              & \multicolumn{2}{c|}{\textbf{Offline LiRA}}            & \multicolumn{2}{c|}{\textbf{Online RMIA}}             & \multicolumn{2}{c}{\textbf{Offline RMIA}}             \\ \cline{2-13} 
\multicolumn{1}{c|}{}                         & \multicolumn{1}{c|}{\textbf{AUC}}        & \textbf{TPR@1\%FPR} & \multicolumn{1}{c|}{\textbf{AUC}}         & \textbf{TPR@1\%FPR} & \multicolumn{1}{c|}{\textbf{AUC}}         & \textbf{TPR@1\%FPR} & \multicolumn{1}{c|}{\textbf{AUC}}        & \textbf{TPR@1\%FPR} & \multicolumn{1}{c|}{\textbf{AUC}}        & \textbf{TPR@1\%FPR} & \multicolumn{1}{c|}{\textbf{AUC}}        & \textbf{TPR@1\%FPR} \\ \hline
Adult                                         & \multicolumn{1}{c|}{49.51 ± 0.45} & 0.82 ± 0.13  & \multicolumn{1}{c|}{49.77 ± 0.28}  & 0.94 ± 0.18  & \multicolumn{1}{c|}{50.9 ± 0.65}   & 1.28 ± 0.2   & \multicolumn{1}{c|}{\textbf{51.05 ± 0.62}} & 1.3 ± 0.25   & \multicolumn{1}{c|}{50.57 ± 0.43} & 1.23 ± 0.17  & \multicolumn{1}{c|}{50.93 ± 0.48} & \textbf{1.42 ± 0.06}  \\
Cirrhosis                                     & \multicolumn{1}{c|}{46.09 ± 7.81} & 0.59 ± 0.58  & \multicolumn{1}{c|}{49.98 ± 6.05}  & 1.25 ± 1.09  & \multicolumn{1}{c|}{\textbf{58.73 ± 3.65}}  & \textbf{4.81 ± 3.29}  & \multicolumn{1}{c|}{58.59 ± 3.63} & 3.52 ± 3.22  & \multicolumn{1}{c|}{56.18 ± 3.56} & 1.39 ± 0.54  & \multicolumn{1}{c|}{58.33 ± 2.58} & 3.38 ± 3.75  \\
Covid                                         & \multicolumn{1}{c|}{48.12 ± 3.69} & 0.76 ± 0.53  & \multicolumn{1}{c|}{46.91 ± 3.18}  & 0.71 ± 0.4   & \multicolumn{1}{c|}{\textbf{48.24 ± 4.08}}  & 1.39 ± 3.1   & \multicolumn{1}{c|}{46.78 ± 2.35} & \textbf{5.56 ± 6.99}  & \multicolumn{1}{c|}{46.85 ± 1.89} & 0.84 ± 0.15  & \multicolumn{1}{c|}{43.13 ± 3.62} & 0.89 ± 0.83  \\
Dropout Success                               & \multicolumn{1}{c|}{46.78 ± 3.74} & 0.59 ± 0.44  & \multicolumn{1}{c|}{47.84 ± 5.94}  & 0.56 ± 0.46  & \multicolumn{1}{c|}{59.14 ± 10.01} & 2.4 ± 2.12   & \multicolumn{1}{c|}{58.78 ± 8.15} & 2.62 ± 1.67  & \multicolumn{1}{c|}{\textbf{60.18 ± 6.82}} & 3.01 ± 1.92  & \multicolumn{1}{c|}{59.62 ± 9.81} & \textbf{5.24 ± 4.6}   \\
Gallstone                                     & \multicolumn{1}{c|}{45.23 ± 8.18} & 0.58 ± 0.53  & \multicolumn{1}{c|}{50.25 ± 4.22}  & 1.7 ± 1.78   & \multicolumn{1}{c|}{\textbf{51.62 ± 2.37}}  & \textbf{2.93 ± 2.64}  & \multicolumn{1}{c|}{49.95 ± 1.42} & 1.2 ± 1.01   & \multicolumn{1}{c|}{49.59 ± 2.09} & 0.94 ± 0.23  & \multicolumn{1}{c|}{50.29 ± 2.07} & 2.43 ± 3.15  \\
Half Million                                  & \multicolumn{1}{c|}{45.74 ± 4.07} & 0.56 ± 0.38  & \multicolumn{1}{c|}{45.17 ± 8.91}  & 0.64 ± 0.51  & \multicolumn{1}{c|}{\textbf{54.2 ± 2.57}}   & 1.17 ± 0.68  & \multicolumn{1}{c|}{51.64 ± 2.97} & \textbf{2.02 ± 0.89}  & \multicolumn{1}{c|}{52.26 ± 2.86} & 1.43 ± 0.35  & \multicolumn{1}{c|}{53.81 ± 2.79} & 1.73 ± 0.47  \\
Locations                                     & \multicolumn{1}{c|}{44.04 ± 8.54} & 0.7 ± 0.83   & \multicolumn{1}{c|}{38.5 ± 10.97}  & 0.4 ± 0.64   & \multicolumn{1}{c|}{\textbf{58.25 ± 9.22}}  & 2.5 ± 2.5    & \multicolumn{1}{c|}{55.68 ± 9.81} & 7.5 ± 9.35   & \multicolumn{1}{c|}{55.14 ± 7.0}  & 1.88 ± 1.66  & \multicolumn{1}{c|}{55.79 ± 11.6} &\textbf{ 8.49 ± 10.36} \\
Purchases-20                                  & \multicolumn{1}{c|}{45.42 ± 2.31} & 0.2 ± 0.37   & \multicolumn{1}{c|}{42.39 ± 7.95}  & 0.33 ± 0.46  & \multicolumn{1}{c|}{\textbf{53.23 ± 2.02}}  & 1.16 ± 0.26  & \multicolumn{1}{c|}{52.63 ± 1.93} & 1.23 ± 0.07  & \multicolumn{1}{c|}{52.75 ± 2.5}  & 1.1 ± 0.29   & \multicolumn{1}{c|}{52.51 ± 1.75} & \textbf{1.33 ± 0.24}  \\
Student Performance                           & \multicolumn{1}{c|}{39.4 ± 8.6}   & 0.59 ± 0.58  & \multicolumn{1}{c|}{49.65 ± 8.35}  & \textbf{4.03 ± 7.08}  & \multicolumn{1}{c|}{\textbf{49.96 ± 1.26}}  & 1.41 ± 0.57  & \multicolumn{1}{c|}{49.42 ± 0.33} & 1.17 ± 0.33  & \multicolumn{1}{c|}{49.78 ± 0.89} & 1.02 ± 0.08  & \multicolumn{1}{c|}{49.56 ± 0.58} & 1.16 ± 0.41  \\ \hline
\end{tabular}%
}
\caption{Performance of MIAs for nine datasets, averaged across six learning models in a centralized paradigm.}
\label{tab:tprfpr}
\end{table*}

\subsection{Target model accuracy results}
To better understand the effects of overfitting (demonstrated through generalization gap), we provide in Table~\ref{tab:target_acc} all the training and testing accuracies for each target model in the different learning paradigms.
\begin{table*}[!ht]
\centering
\scriptsize
\resizebox{\textwidth}{!}{%
\renewcommand{\arraystretch}{1}
\begin{tabular}{c|c|ccccccccc}
\hline
\multirow{2}{*}{\textbf{Train Mode}}                                          & \multirow{2}{*}{\textbf{Model}} & \multicolumn{9}{c}{\textbf{Dataset}}                                                                                                     \\ \cline{3-11} 
                                                                              &                                 & Adult       & Cirrhosis   & Covid       & Dropout Success & Gallstone  & Half Million & Locations   & Purchases-20 & Student Performance \\ \hline
\multirow{6}{*}{\begin{tabular}[c]{@{}c@{}}Central No\\ Defense\end{tabular}} & NN                              & 85.57 / 84.33 & 81.41 / 69.81 & 81.42 / 70.36 & 86.98 / 74.14     & 94.12 / 85.0 & 72.07 / 63.02  & 100.0 / 59.33 & 100.0 / 90.87  & 79.42 / 55.56         \\
                                                                              & NB                              & 81.34 / 82.43 & 76.92 / 64.15 & 68.33 / 70.06 & 69.14 / 67.99     & 73.11 / 62.5 & 50.57 / 49.95  & 70.29 / 38.92 & 75.95 / 72.59  & 60.49 / 50.62         \\
                                                                              & DT                              & 85.83 / 85.32 & 94.23 / 62.26 & 78.72 / 66.17 & 79.57 / 71.61     & 94.96 / 72.5 & 68.37 / 53.13  & 50.37 / 27.43 & 52.36 / 39.17  & 93.0 / 40.74          \\
                                                                              & RF                              & 86.99 / 86.17 & 93.59 / 79.25 & 91.01 / 70.36 & 88.43 / 75.77     & 100.0 / 75.0 & 81.82 / 62.01  & 97.87 / 52.63 & 87.93 / 61.1   & 97.94 / 61.73         \\
                                                                              & XGB                             & 90.90 / 86.33 & 99.36 / 69.81 & 99.7 / 68.86  & 100.0 / 74.5      & 100.0 / 80.0 & 100.0 / 66.45  & 100.0 / 58.69 & 100.0 / 83.55  & 100.0 / 60.49         \\
                                                                              & LR                              & 81.9 / 83.17  & 83.33 / 73.58 & 75.22 / 70.66 & 77.76 / 75.77     & 100.0 / 85.0 & 62.02 / 60.9   & 100.0 / 55.34 & 100.0 / 88.28  & 71.6 / 55.56          \\ \hline
\multirow{6}{*}{\begin{tabular}[c]{@{}c@{}}Central \\ Defense\end{tabular}}   & NN                              & 83.95 / 84.63 & 71.79 / 64.15 & 68.83 / 68.56 & 74.38 / 71.97     & 60.5 / 67.5  & 56.34 / 54.69  & 82.06 / 54.86 & 98.48 / 91.13  & 49.79 / 46.91         \\
                                                                              & NB                              & 81.4 / 82.46  & 74.36 / 66.04 & 68.33 / 70.06 & 69.26 / 67.99     & 73.11 / 62.5 & 50.66 / 50.1   & 70.29 / 38.92 & 76.22 / 72.92  & 60.49 / 50.62         \\
                                                                              & DT                              & 85.44 / 85.35 & 82.69 / 67.92 & 74.33 / 71.86 & 79.39 / 71.61     & 84.03 / 60.0 & 61.47 / 54.74  & 36.37 / 26.0  & 42.47 / 38.42  & 70.78 / 58.02         \\
                                                                              & RF                              & 85.94 / 86.06 & 79.49 / 75.47 & 80.32 / 70.06 & 87.76 / 74.5      & 100.0 / 80.0 & 69.46 / 60.85  & 67.52 / 44.5  & 62.22 / 53.5   & 95.88 / 61.73         \\
                                                                              & XGB                             & 86.8 / 86.2   & 78.79 / 73.58 & 76.0 / 68.86  & 85.46 / 75.95     & 76.24 / 65.0 & 98.93 / 65.79  & 95.61 / 63.16 & 100.0 / 87.81  & 71.6 / 55.56          \\
                                                                              & LR                              & 81.9 / 83.2   & 80.77 / 73.58 & 75.52 / 70.36 & 77.58 / 75.77     & 92.44 / 85.0 & 61.77 / 60.24  & 99.52 / 58.69 & 99.99 / 88.64  & 71.6 / 59.26          \\ \hline
\multirow{3}{*}{\begin{tabular}[c]{@{}c@{}}Federated\\ Learning\end{tabular}} & NN                              & 84.31 / 84.77 & 81.41 / 71.7  & 71.53 / 69.16 & 77.64 / 74.14     & 65.55 / 75.0 & 60.04 / 58.32  & 96.01 / 60.93 & 100.0 / 91.56  & 55.14 / 54.32         \\
                                                                              & XGB                             & 86.72 / 85.92 & 77.56 / 69.81 & 72.63 / 59.88 & 78.84 / 69.26     & 75.63 / 62.5 & 42.26 / 37.59  & 23.06 / 20.41 & 38.81 / 31.16  & 59.26 / 54.32         \\
                                                                              & LR                              & 81.75 / 82.71 & 81.41 / 75.47 & 75.12 / 70.96 & 77.58 / 75.41     & 78.15 / 82.5 & 55.13 / 54.69  & 98.83 / 60.13 & 84.01 / 81.74  & 66.26 / 53.09         \\ \hline
\end{tabular}%
}

\caption{Target model accuracy (training / testing) results for each dataset and training environment.}
\label{tab:target_acc}
\end{table*}

\subsection{MIAs accuracy on XGBoost}
\begin{figure}[!ht]
\centerline{
\includegraphics[width=\columnwidth]{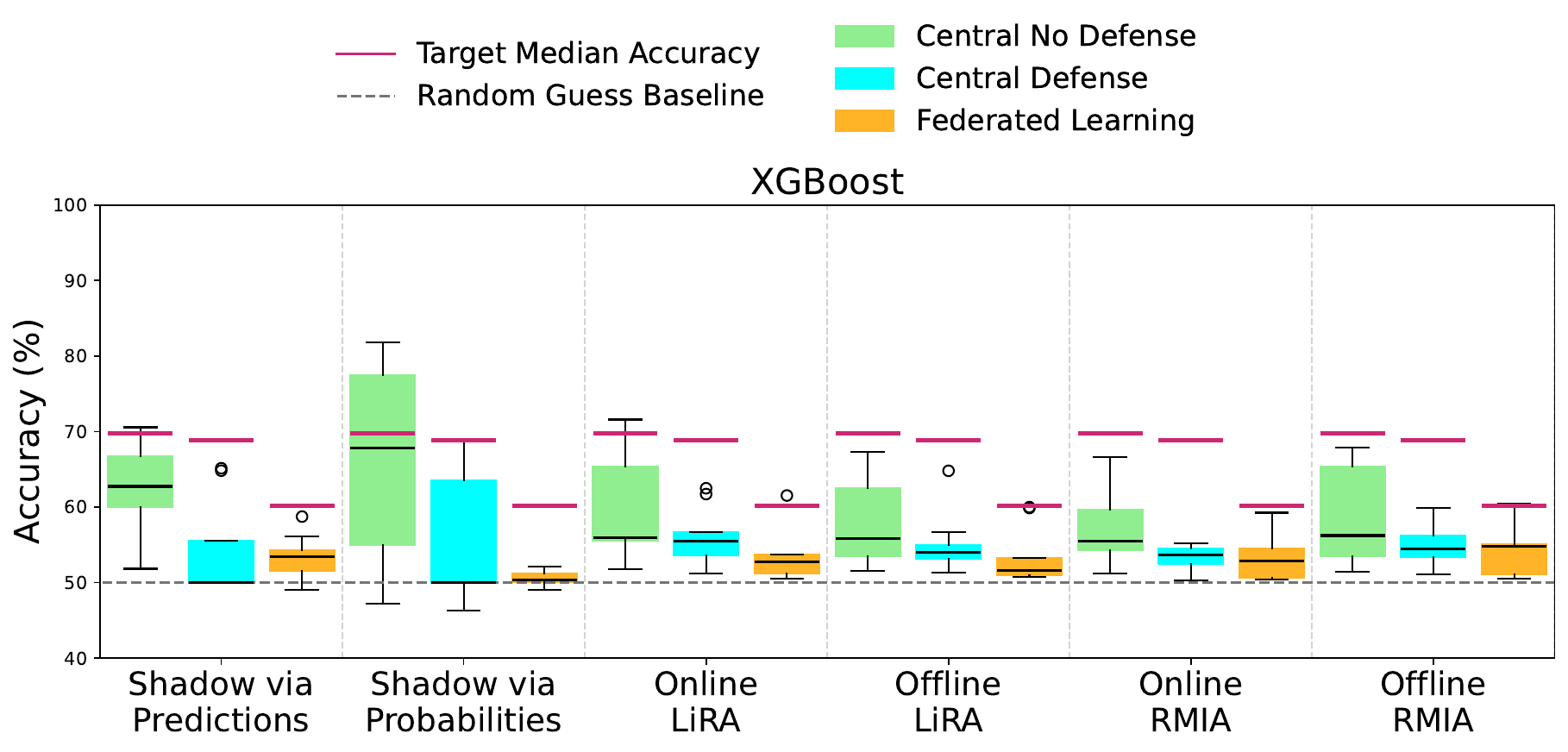}
}
\caption{Comparison of MIAs efficacy on XGBoost considering accuracy as the evaluation metric.}
\label{fig:xgb_acc}
\end{figure}

Although we use only a single shadow model, we can verify in Figure~\ref{fig:xgb_acc} that shadow model-based attacks achieve superior performance when evaluated using accuracy. In particular, attacks with shadow via probabilities demonstrate a clear advantage, with a median accuracy close to 70\%, highlighting the effectiveness of these attacks under averaged metrics, even with one shadow model. However, when evaluated using AUC, their performance drops substantially, indicating that accuracy alone can overestimate attack success and underscoring the importance of emphasizing positive predictions over non-membership predictions for a more reliable assessment.

\end{document}